\documentclass{article}

\usepackage{microtype}
\usepackage{graphicx}
\usepackage{subfigure}
\usepackage{booktabs} 

\usepackage{hyperref}

\usepackage[accepted]{icml2026}

\usepackage{amsmath}
\usepackage{amssymb}
\usepackage{mathtools}
\usepackage{amsthm}
\usepackage{cancel}

\usepackage[capitalize,noabbrev]{cleveref}
\usepackage{etoolbox}

\theoremstyle{plain}
\newtheorem{theorem}{Theorem}[section]

\theoremstyle{definition}
\newtheorem{example}{Example}
\theoremstyle{remark}
\newtheorem{remark}[theorem]{Remark}

\newcommand{\myparagraph}[1]{\textbf{#1} \hspace{0.5em}}

\usepackage{mdframed}

\newtheorem*{lemma*}{Lemma}
\newtheorem*{proposition*}{Proposition}
\newtheorem*{corollary*}{Corollary}
\newtheorem*{conjecture*}{Conjecture}

\newmdtheoremenv[
  backgroundcolor=blue!10,   
  linecolor=blue!50,         
  innertopmargin=8pt,
  innerbottommargin=2pt,
  skipabove=2pt,
  skipbelow=2pt
]{definition2}{Definition}

\newmdtheoremenv[
  backgroundcolor=red!10,   
  linecolor=red!50,         
  innertopmargin=8pt,
  innerbottommargin=2pt,
  skipabove=2pt,
  skipbelow=2pt
]{definition3}{Symmetry}

\newmdtheoremenv[
  backgroundcolor=green!10,   
  linecolor=green!50,         
  innertopmargin=8pt,
  innerbottommargin=2pt,
  skipabove=2pt,
  skipbelow=2pt
]{definition4}{Conclusion}

\usepackage{mathtools}

\usepackage{tikz}
\usetikzlibrary{arrows.meta, positioning, quotes, fit, shapes, backgrounds, calc}

\tikzstyle{none}=[]
\tikzstyle{white dot}=[fill=white, draw=black, shape=circle, text width=.1cm, inner sep=.03cm]
\tikzstyle{black dot}=[fill=black, draw=white, shape=circle, draw=white, text width=.1cm, inner sep=.03cm]
\tikzstyle{red dot}=[fill={rgb,255: red,163; green,0; blue,2}, draw=black, shape=circle]
\tikzstyle{medium box}=[fill=white, draw=black, shape=rectangle, minimum width=2cm, minimum height=1.cm, text=black]
\tikzstyle{small box}=[fill=white, draw=black, shape=rectangle, minimum width=.5cm, minimum height=.5cm, text=black]
\tikzstyle{triangle}=[fill=white, draw=black, regular polygon, regular polygon sides=3, shape border rotate=90, minimum width=1.cm, text width=.3cm, inner sep=.0cm, align=center, anchor=center, text=black]
\tikzstyle{triangleflip}=[fill=white, draw=black, regular polygon, regular polygon sides=3, shape border rotate=-90, minimum width=1.cm, text width=.3cm, inner sep=.0cm, align=center, anchor=center, text=black]

\tikzstyle{new edge style 0}=[-, fill=none, draw={rgb,255: red,0; green,172; blue,187}]
\tikzstyle{thick edge}=[-, line width=.03cm]
\tikzstyle{norm}=[-, dashed, draw={rgb,255: red,46; green,126; blue,255}]
\tikzstyle{amax}=[-, dashed, draw={rgb,255: red,255; green,128; blue,55}]

\pgfdeclarelayer{nodelayer}
\pgfdeclarelayer{edgelayer}
\pgfsetlayers{nodelayer,background,main,edgelayer}

\tikzset{
    arrowstyledash/.style={->, thick, rounded corners, dash pattern=on 2pt off .6pt, gray},
}

\newcommand{\probabilityDistribution}{
\begin{tikzpicture}
	\begin{pgfonlayer}{nodelayer}
		\node [style=small box] (132) at (2.925, 3.25) {$\pdist_{\rv{X}}$};
		\node [style=none] (136) at (3.875, 3.25) {};
		\node [style=none] (138) at (4.175, 3.25) {$\rv{X}$};
		\node [style=none] (139) at (1.175, 3.25) {$\pdist_\rv{X}$};
		\node [style=none] (140) at (1.925, 3.25) {$=$};
	\end{pgfonlayer}
	\begin{pgfonlayer}{edgelayer}
		\draw (132) to (136.center);
	\end{pgfonlayer}
\end{tikzpicture}

}

\newcommand{\parallelProbabilisticComposition}{
\begin{tikzpicture}
	\begin{pgfonlayer}{nodelayer}
		\node [style=small box] (0) at (9.675, 3.25) {$\pdist_{\rv{X},\rv{Y}}$};
		\node [style=none] (1) at (11.125, 3.5) {};
		\node [style=none] (2) at (11.425, 3.5) {$\rv{X}$};
		\node [style=none] (3) at (11.125, 3) {};
		\node [style=none] (4) at (11.425, 3) {$\rv{Y}$};
		\node [style=none] (5) at (8.675, 3.25) {$=$};
		\node [style=small box] (6) at (6.425, 3.625) {$\pdist_{\rv{X}}$};
		\node [style=none] (7) at (7.625, 3.625) {};
		\node [style=none] (8) at (7.925, 3.625) {$\rv{X}$};
		\node [style=none] (9) at (7.625, 2.875) {};
		\node [style=none] (10) at (7.925, 2.875) {$\rv{Y}$};
		\node [style=small box] (11) at (6.425, 2.875) {$\pdist_{\rv{Y}}$};
		\node [style=none] (12) at (12.175, 3.25) {$=$};
		\node [style=none] (13) at (13.175, 3.25) {$\pdist_{\rv{X},\rv{Y}}$};
		\node [style=none] (14) at (4.175, 3.25) {$\pdist_{\rv{X}} \otimes \pdist_{\rv{Y}}$};
		\node [style=none] (15) at (5.425, 3.25) {$=$};
	\end{pgfonlayer}
	\begin{pgfonlayer}{edgelayer}
		\draw [in=180, out=15, looseness=1.25] (0) to (1.center);
		\draw [in=-180, out=-15, looseness=1.25] (0) to (3.center);
		\draw (6) to (7.center);
		\draw (11) to (9.center);
	\end{pgfonlayer}
\end{tikzpicture}

}

\newcommand{\copyProbabilistic}{
\begin{tikzpicture}
	\begin{pgfonlayer}{nodelayer}
		\node [style=none] (22) at (4.675, 1.75) {$\rv{X}$};
		\node [style=none] (23) at (5.925, 2.25) {};
		\node [style=none] (24) at (5.925, 1.25) {};
		\node [style=white dot] (25) at (5.625, 1.75) {};
		\node [style=none] (26) at (5.175, 1.75) {};
		\node [style=none] (27) at (6.425, 2.25) {$\rv{X}$};
		\node [style=none] (28) at (6.425, 1.25) {$\rv{X}$};
	\end{pgfonlayer}
	\begin{pgfonlayer}{edgelayer}
		\draw [in=90, out=180] (23.center) to (25);
		\draw [in=-90, out=-180] (24.center) to (25);
		\draw (26.center) to (25);
	\end{pgfonlayer}
\end{tikzpicture}

}

\newcommand{\sequentialProbabilisticComposition}{
\begin{tikzpicture}
	\begin{pgfonlayer}{nodelayer}
		\node [style=small box] (0) at (6.175, 3.25) {$\pdist_{\rv{Y} \mid \rv{X}}$};
		\node [style=none] (1) at (7.1, 3.25) {};
		\node [style=none] (2) at (7.4, 3.25) {$\rv{Y}$};
		\node [style=small box] (3) at (3.925, 3.25) {$\pdist_{\rv{X}}$};
		\node [style=none] (4) at (5.3, 3.75) {};
		\node [style=white dot] (5) at (5, 3.25) {};
		\node [style=none] (6) at (7.1, 3.75) {};
		\node [style=none] (7) at (7.4, 3.75) {$\rv{X}$};
		\node [style=none] (13) at (3.15, 3.25) {$=$};
		\node [style=small box] (14) at (9.175, 3.25) {$\pdist_{\rv{X},\rv{Y}}$};
		\node [style=none] (15) at (10.625, 3.5) {};
		\node [style=none] (16) at (10.925, 3.5) {$\rv{X}$};
		\node [style=none] (17) at (10.625, 3) {};
		\node [style=none] (18) at (10.925, 3) {$\rv{Y}$};
		\node [style=none] (19) at (8.175, 3.25) {$=$};
		\node [style=none] (20) at (11.675, 3.25) {$=$};
		\node [style=none] (21) at (12.425, 3.25) {$\pdist_{\rv{X},\rv{Y}}$};
		\node [style=none] (22) at (2.175, 3.25) {$\pdist_{\rv{X}} \circ \pdist_{\rv{Y}\mid\rv{X}}$};
	\end{pgfonlayer}
	\begin{pgfonlayer}{edgelayer}
		\draw (0) to (1.center);
		\draw [in=90, out=180] (4.center) to (5);
		\draw (3) to (5);
		\draw (5) to (0);
		\draw (4.center) to (6.center);
		\draw [in=180, out=15, looseness=1.25] (14) to (15.center);
		\draw [in=-180, out=-15, looseness=1.25] (14) to (17.center);
	\end{pgfonlayer}
\end{tikzpicture}

}

\newcommand{\marginalProbabilisticInference}{
\begin{tikzpicture}
	\begin{pgfonlayer}{nodelayer}
		\node [style=small box] (0) at (2.75, 2.25) {$\pdist_{\rv{Y} \mid \rv{X}}$};
		\node [style=none] (18) at (3.925, 2.25) {};
		\node [style=none] (19) at (4.225, 2.25) {$Y$};
		\node [style=none] (21) at (1.7, 2.25) {};
		\node [style=none] (22) at (1.7, 1.25) {};
		\node [style=white dot] (23) at (1.4, 1.75) {};
		\node [style=none] (24) at (3.825, 1.25) {};
		\node [style=none] (25) at (3.825, 1.25) {};
		\node [style=none] (26) at (3.825, 1) {};
		\node [style=none] (27) at (3.825, 1.5) {};
		\node [style=none] (28) at (3.925, 1.1) {};
		\node [style=none] (29) at (3.925, 1.4) {};
		\node [style=none] (30) at (4.025, 1.2) {};
		\node [style=none] (31) at (4.025, 1.3) {};
		\node [style=small box] (33) at (0.15, 1.75) {$\pdist_{\rv{X}}$};
		\node [style=none] (34) at (4.675, 1.75) {$=$};
		\node [style=small box] (35) at (6.775, 1.75) {$\pdist_{\rv{Y} \mid \rv{X}}$};
		\node [style=none] (36) at (7.95, 1.75) {};
		\node [style=none] (37) at (8.25, 1.75) {$Y$};
		\node [style=small box] (40) at (5.425, 1.75) {$\pdist_{\rv{X}}$};
	\end{pgfonlayer}
	\begin{pgfonlayer}{edgelayer}
		\draw (0) to (18.center);
		\draw [in=90, out=180] (21.center) to (23);
		\draw [in=-90, out=-180] (22.center) to (23);
		\draw (21.center) to (0);
		\draw [style=thick edge] (27.center) to (26.center);
		\draw [style=thick edge] (29.center) to (28.center);
		\draw [style=thick edge] (31.center) to (30.center);
		\draw (22.center) to (25.center);
		\draw (33) to (23);
		\draw (35) to (36.center);
		\draw (40) to (35);
	\end{pgfonlayer}
\end{tikzpicture}

}

\newcommand{\parametricModel}{
\begin{tikzpicture}
	\begin{pgfonlayer}{nodelayer}
		\node [style=small box] (5) at (7.725, 2.625) {$\pdist_{\rv{X}}$};
		\node [style=small box] (10) at (9.725, 3.25) {$\pdist_{\rv{Y} \mid \rv{X}; \Theta}$};
		\node [style=none] (11) at (10.975, 3.25) {};
		\node [style=none] (12) at (11.275, 3.25) {$\rv{Y}$};
		\node [style=white dot] (14) at (8.525, 2.625) {};
		\node [style=none] (17) at (10.975, 2.625) {};
		\node [style=none] (18) at (11.275, 2.625) {$\rv{X}$};
	\end{pgfonlayer}
	\begin{pgfonlayer}{edgelayer}
		\draw (10) to (11.center);
		\draw (5) to (14);
		\draw [in=-165, out=15] (14) to (10);
		\draw (14) to (17.center);
	\end{pgfonlayer}
\end{tikzpicture}

}

\newcommand{\posteriorParametricInference}{
\begin{tikzpicture}
	\begin{pgfonlayer}{nodelayer}
		\node [style=none] (33) at (-0.75, 1.25) {};
		\node [style=none] (34) at (-0.75, -1.35) {};
		\node [style=none] (35) at (3.525, 1.25) {};
		\node [style=none] (36) at (3.525, -1.35) {};
		\node [style=small box] (40) at (-0.275, 0.5) {$\pdist_{\Theta}$};
		\node [style=triangle] (41) at (-1.275, -0.2) {$x$};
		\node [style=small box] (42) at (1.725, 0.5) {$\pdist_{\rv{Y} \mid \rv{X}; \Theta}$};
		\node [style=white dot] (45) at (0.525, 0.5) {};
		\node [style=none] (47) at (3.975, 1) {};
		\node [style=none] (48) at (4.275, 1) {$\Theta$};
		\node [style=none] (51) at (1.225, 1) {};
		\node [style=none] (52) at (2.725, 0.5) {};
		\node [style=none] (53) at (2.725, -1.15) {};
		\node [style=none] (54) at (3.375, -0.325) {};
		\node [style=triangle] (55) at (-1.275, -1.15) {$y$};
	\end{pgfonlayer}
	\begin{pgfonlayer}{edgelayer}
		\draw [style=norm] (33.center) to (35.center);
		\draw [style=norm] (35.center) to (36.center);
		\draw [style=norm] (33.center) to (34.center);
		\draw [style=norm] (34.center) to (36.center);
		\draw (40) to (45);
		\draw (45) to (42);
		\draw (51.center) to (47.center);
		\draw [in=180, out=90] (45) to (51.center);
		\draw [in=90, out=0] (52.center) to (54.center);
		\draw [in=-90, out=0] (53.center) to (54.center);
		\draw (42) to (52.center);
		\draw (55) to (53.center);
		\draw [in=-165, out=0] (41) to (42);
	\end{pgfonlayer}
\end{tikzpicture}

}

\newcommand{\mapInference}{
\begin{tikzpicture}
	\begin{pgfonlayer}{nodelayer}
		\node [style=small box] (4) at (1.5, 0.5) {$\pdist_{\Theta}$};
		\node [style=triangle] (5) at (0.5, -0.2) {$x$};
		\node [style=small box] (6) at (3.5, 0.5) {$\pdist_{\rv{Y} \mid \rv{X}; \Theta}$};
		\node [style=white dot] (7) at (2.3, 0.5) {};
		\node [style=triangleflip] (9) at (6.05, 1) {$\theta$};
		\node [style=none] (10) at (3, 1) {};
		\node [style=none] (11) at (4.5, 0.5) {};
		\node [style=none] (12) at (4.5, -1.15) {};
		\node [style=none] (13) at (5.15, -0.325) {};
		\node [style=triangle] (14) at (0.5, -1.15) {$y$};
	\end{pgfonlayer}
	\begin{pgfonlayer}{edgelayer}
		\draw (4) to (7);
		\draw (7) to (6);
		\draw [in=180, out=90] (7) to (10.center);
		\draw [in=90, out=0] (11.center) to (13.center);
		\draw [in=-90, out=0] (12.center) to (13.center);
		\draw (6) to (11.center);
		\draw (14) to (12.center);
		\draw [in=-165, out=0] (5) to (6);
		\draw (10.center) to (9);
	\end{pgfonlayer}
\end{tikzpicture}

}

\newcommand{\mleInference}{
\begin{tikzpicture}
	\begin{pgfonlayer}{nodelayer}
		\node [style=triangle] (5) at (0.5, -0.2) {$x$};
		\node [style=small box] (6) at (3.5, 0.5) {$\pdist_{\rv{Y} \mid \rv{X}; \Theta}$};
		\node [style=triangleflip] (9) at (6.05, 1) {$\theta$};
		\node [style=none] (10) at (3.25, 1) {};
		\node [style=none] (11) at (4.5, 0.5) {};
		\node [style=none] (12) at (4.5, -1.15) {};
		\node [style=none] (13) at (5.15, -0.325) {};
		\node [style=triangle] (14) at (0.5, -1.15) {$y$};
		\node [style=none] (15) at (2.55, 0.75) {};
	\end{pgfonlayer}
	\begin{pgfonlayer}{edgelayer}
		\draw [in=90, out=0] (11.center) to (13.center);
		\draw [in=-90, out=0] (12.center) to (13.center);
		\draw (6) to (11.center);
		\draw (14) to (12.center);
		\draw [in=-165, out=0] (5) to (6);
		\draw [in=-90, out=180, looseness=0.75] (6) to (15.center);
		\draw [in=180, out=90] (15.center) to (10.center);
		\draw (10.center) to (9);
	\end{pgfonlayer}
\end{tikzpicture}

}

\newcommand{\manifoldReparametrisation}{
\begin{tikzpicture}
	\begin{pgfonlayer}{nodelayer}
		\node [style=small box] (0) at (2.675, 3.25) {$\pdist_{\rv{X}}$};
		\node [style=small box] (1) at (4.175, 3.25) {$\pdist_{\rv{Z} \mid \rv{X}}$};
		\node [style=small box] (2) at (5.925, 3.25) {$\pdist_{\rv{Y} \mid \rv{Z}}$};
		\node [style=none] (3) at (7, 3.25) {};
		\node [style=none] (4) at (7.25, 3.25) {$\rv{Y}$};
		\node [style=small box] (5) at (-4, 3.25) {$\pdist_{\rv{X}}$};
		\node [style=small box] (6) at (-2.5, 3.25) {$\pdist_{\rv{Y} \mid \rv{X}}$};
		\node [style=none] (7) at (0.5, 3.4) {$\xrightarrow{\text{manifold assumption}}$};
		\node [style=none] (8) at (-1.5, 3.25) {};
		\node [style=none] (9) at (-1.25, 3.25) {$\rv{Y}$};
	\end{pgfonlayer}
	\begin{pgfonlayer}{edgelayer}
		\draw (0) to (1);
		\draw (1) to (2);
		\draw (2) to (3.center);
		\draw (5) to (6);
		\draw (6) to (8.center);
	\end{pgfonlayer}
\end{tikzpicture}

}

\newcommand{\conceptMembership}{
\begin{tikzpicture}
	\begin{pgfonlayer}{nodelayer}
		\node [style=small box] (1) at (4.175, 3.25) {$\pdist_{\rv{C} \mid \rv{Z}; \rv{T}}$};
		\node [style=none] (3) at (5.25, 3.25) {};
		\node [style=none] (4) at (5.5, 3.25) {$\rv{C}$};
		\node [style=none] (5) at (3, 3.25) {};
		\node [style=none] (6) at (2.75, 3.25) {$\rv{Z}$};
	\end{pgfonlayer}
	\begin{pgfonlayer}{edgelayer}
		\draw (1) to (3.center);
		\draw (5.center) to (1);
	\end{pgfonlayer}
\end{tikzpicture}

}

\newcommand{\conceptToConcept}{
\begin{tikzpicture}
	\begin{pgfonlayer}{nodelayer}
		\node [style=small box] (0) at (4.175, 3.25) {$\pdist_{\rv{C} \mid \rv{C}; \rv{T}}$};
		\node [style=none] (1) at (5.25, 3.25) {};
		\node [style=none] (2) at (5.5, 3.25) {$\rv{C}$};
		\node [style=none] (3) at (3, 3.25) {};
		\node [style=none] (4) at (2.75, 3.25) {$\rv{C}$};
	\end{pgfonlayer}
	\begin{pgfonlayer}{edgelayer}
		\draw (0) to (1.center);
		\draw (3.center) to (0);
	\end{pgfonlayer}
\end{tikzpicture}

}

\newcommand{\taskPredictor}{
\begin{tikzpicture}
	\begin{pgfonlayer}{nodelayer}
		\node [style=small box] (0) at (4.175, 3.25) {$\pdist_{\rv{Y} \mid \rv{C}}$};
		\node [style=none] (1) at (5.25, 3.25) {};
		\node [style=none] (2) at (5.5, 3.25) {$\rv{Y}$};
		\node [style=none] (3) at (3, 3.25) {};
		\node [style=none] (4) at (2.75, 3.25) {$\rv{C}$};
	\end{pgfonlayer}
	\begin{pgfonlayer}{edgelayer}
		\draw (0) to (1.center);
		\draw (3.center) to (0);
	\end{pgfonlayer}
\end{tikzpicture}

}

\newcommand{\compositions}{
\begin{tikzpicture}
	\begin{pgfonlayer}{nodelayer}
		\node [style=small box] (1) at (-3.45, 3.25) {$f$};
		\node [style=none] (13) at (-4.15, 3.25) {};
		\node [style=none] (22) at (-3.7, 3.25) {};
		\node [style=none] (23) at (-3.2, 3.25) {};
		\node [style=none] (24) at (-2.75, 3.25) {};
		\node [style=none] (27) at (-2.425, 3.25) {$\circ$};
		\node [style=none] (35) at (-0.325, 3.25) {$=$};
		\node [style=small box] (51) at (-1.4, 3.25) {$g$};
		\node [style=none] (52) at (-2.1, 3.25) {};
		\node [style=none] (53) at (-1.65, 3.25) {};
		\node [style=none] (54) at (-1.15, 3.25) {};
		\node [style=none] (55) at (-0.7, 3.25) {};
		\node [style=small box] (58) at (0.775, 3.25) {$g$};
		\node [style=none] (59) at (0.075, 3.25) {};
		\node [style=none] (60) at (0.525, 3.25) {};
		\node [style=none] (61) at (1.025, 3.25) {};
		\node [style=none] (62) at (1.725, 3.25) {};
		\node [style=small box] (66) at (1.975, 3.25) {$f$};
		\node [style=none] (69) at (2.225, 3.25) {};
		\node [style=none] (70) at (2.675, 3.25) {};
		\node [style=small box] (71) at (4.525, 3.25) {$f$};
		\node [style=none] (72) at (3.825, 3.25) {};
		\node [style=none] (73) at (4.275, 3.25) {};
		\node [style=none] (74) at (4.775, 3.25) {};
		\node [style=none] (75) at (5.225, 3.25) {};
		\node [style=none] (76) at (5.55, 3.25) {$\otimes$};
		\node [style=none] (77) at (7.65, 3.25) {$=$};
		\node [style=small box] (78) at (6.575, 3.25) {$g$};
		\node [style=none] (79) at (5.875, 3.25) {};
		\node [style=none] (80) at (6.325, 3.25) {};
		\node [style=none] (81) at (6.825, 3.25) {};
		\node [style=none] (82) at (7.275, 3.25) {};
		\node [style=small box] (83) at (8.775, 3.625) {$f$};
		\node [style=none] (84) at (8.075, 3.625) {};
		\node [style=none] (85) at (8.525, 3.625) {};
		\node [style=none] (86) at (9.025, 3.625) {};
		\node [style=none] (87) at (9.475, 3.625) {};
		\node [style=small box] (88) at (8.775, 2.875) {$g$};
		\node [style=none] (89) at (8.075, 2.875) {};
		\node [style=none] (90) at (8.525, 2.875) {};
		\node [style=none] (91) at (9.025, 2.875) {};
		\node [style=none] (92) at (9.475, 2.875) {};
		\node [style=none] (93) at (-0.775, 2.25) {Sequential};
		\node [style=none] (94) at (6.725, 2.25) {Parallel};
	\end{pgfonlayer}
	\begin{pgfonlayer}{edgelayer}
		\draw (13.center) to (22.center);
		\draw (23.center) to (24.center);
		\draw (52.center) to (53.center);
		\draw (54.center) to (55.center);
		\draw (59.center) to (60.center);
		\draw (61.center) to (62.center);
		\draw (69.center) to (70.center);
		\draw (72.center) to (73.center);
		\draw (74.center) to (75.center);
		\draw (79.center) to (80.center);
		\draw (81.center) to (82.center);
		\draw (84.center) to (85.center);
		\draw (86.center) to (87.center);
		\draw (89.center) to (90.center);
		\draw (91.center) to (92.center);
	\end{pgfonlayer}
\end{tikzpicture}

}

\newcommand{\interpretableModel}{
\begin{tikzpicture}
	\begin{pgfonlayer}{nodelayer}
		\node [style=small box] (0) at (5.925, 3.25) {$\pdist_{\rv{C}_\rv{T} \mid \rv{X}; \Theta_c}$};
		\node [style=small box] (1) at (2.475, 2.925) {$\pdist_{\rv{X}}$};
		\node [style=small box] (2) at (4, 3.7) {$\pdist_{\Theta_c}$};
		\node [style=none] (7) at (5.075, 4.475) {};
		\node [style=none] (14) at (11.65, 4.475) {};
		\node [style=none] (15) at (11.975, 4.475) {$\Theta_c$};
		\node [style=small box] (19) at (9.675, 3.25) {$\pdist_{\rv{Y}_\rv{T} \mid \rv{C}_\rv{T}; \Theta_y}$};
		\node [style=small box] (21) at (7.5, 3.7) {$\pdist_{\Theta_y}$};
		\node [style=none] (26) at (8.575, 4.1) {};
		\node [style=none] (33) at (11.65, 4.1) {};
		\node [style=none] (34) at (11.975, 4.1) {$\Theta_y$};
		\node [style=white dot] (35) at (4.725, 3.7) {};
		\node [style=white dot] (36) at (8.225, 3.7) {};
		\node [style=white dot] (37) at (7.225, 3.25) {};
		\node [style=none] (38) at (11.65, 3.25) {};
		\node [style=none] (39) at (11.975, 3.25) {$\rv{Y}_\rv{T}$};
		\node [style=none] (40) at (11.65, 2.8) {};
		\node [style=none] (41) at (11.975, 2.8) {$\rv{C}_\rv{T}$};
		\node [style=none] (43) at (7.65, 2.8) {};
		\node [style=white dot] (44) at (3.225, 2.925) {};
		\node [style=none] (45) at (3.65, 2.425) {};
		\node [style=none] (46) at (11.65, 2.425) {};
		\node [style=none] (47) at (11.975, 2.425) {$\rv{X}$};
	\end{pgfonlayer}
	\begin{pgfonlayer}{edgelayer}
		\draw (7.center) to (14.center);
		\draw (26.center) to (33.center);
		\draw (2) to (35);
		\draw [in=165, out=0, looseness=1.25] (35) to (0);
		\draw (21) to (36);
		\draw [in=165, out=0, looseness=1.25] (36) to (19);
		\draw (0) to (37);
		\draw (37) to (19);
		\draw (43.center) to (40.center);
		\draw [in=-180, out=-90, looseness=1.25] (37) to (43.center);
		\draw [in=90, out=-180, looseness=1.25] (26.center) to (36);
		\draw [in=90, out=180] (7.center) to (35);
		\draw [in=-180, out=-90, looseness=1.25] (44) to (45.center);
		\draw [in=-180, out=0] (44) to (0);
		\draw (1) to (44);
		\draw (45.center) to (46.center);
		\draw (19) to (38.center);
	\end{pgfonlayer}
\end{tikzpicture}

}

\newcommand{\parametricConceptProcess}{
\begin{tikzpicture}
	\begin{pgfonlayer}{nodelayer}
		\node [style=small box] (26) at (5.525, 6) {$\pdist_{\rv{C} \mid \text{pa}(\rv{C}); \Theta}$};
		\node [style=none] (27) at (2.95, 5.5) {$\text{pa}(\rv{C})$};
		\node [style=small box] (28) at (3.2, 6.5) {$\pdist_{\Theta}$};
		\node [style=none] (29) at (3.45, 5.5) {};
		\node [style=none] (30) at (6.95, 6) {$\rv{C}$};
		\node [style=none] (31) at (6.7, 6) {};
		\node [style=none] (32) at (4.55, 5.875) {};
		\node [style=none] (33) at (4.55, 6.125) {};
	\end{pgfonlayer}
	\begin{pgfonlayer}{edgelayer}
		\draw (26) to (31.center);
		\draw [in=180, out=0] (29.center) to (32.center);
		\draw [in=180, out=0, looseness=1.25] (28) to (33.center);
	\end{pgfonlayer}
\end{tikzpicture}

}

\newcommand{\conceptAlignment}{
\begin{tikzpicture}
	\begin{pgfonlayer}{nodelayer}
		\node [style=small box] (1) at (4.075, 3.25) {$\pdist_{\rv{C} \mid \text{pa}(\rv{C}); \Theta}$};
		\node [style=none] (6) at (0, 3.125) {$\text{pa}(\s{C})$};
		\node [style=small box] (7) at (1.25, 3.825) {$\pdist_{\Theta}$};
		\node [style=none] (10) at (2.55, 4.275) {};
		\node [style=none] (11) at (2.55, 3.375) {};
		\node [style=white dot] (12) at (2.25, 3.825) {};
		\node [style=none] (13) at (6, 4.275) {};
		\node [style=none] (14) at (6.25, 4.275) {$\Theta$};
		\node [style=none] (15) at (0.5, 3.125) {};
		\node [style=none] (16) at (0.75, 4.5) {};
		\node [style=none] (17) at (0.75, 2) {};
		\node [style=none] (18) at (5.75, 4.5) {};
		\node [style=none] (19) at (5.75, 2) {};
		\node [style=triangle] (20) at (0.1, 2.25) {$c$};
		\node [style=none] (21) at (5, 3.25) {};
		\node [style=none] (22) at (5, 2.25) {};
		\node [style=none] (23) at (5.575, 2.75) {};
		\node [style=none] (24) at (3.2, 3.125) {};
		\node [style=none] (25) at (3.2, 3.375) {};
	\end{pgfonlayer}
	\begin{pgfonlayer}{edgelayer}
		\draw [in=90, out=180] (10.center) to (12);
		\draw [in=-90, out=-180] (11.center) to (12);
		\draw (7) to (12);
		\draw (10.center) to (13.center);
		\draw [style=norm] (16.center) to (17.center);
		\draw [style=norm] (18.center) to (19.center);
		\draw [style=norm] (17.center) to (19.center);
		\draw [style=norm] (16.center) to (18.center);
		\draw [in=90, out=0] (21.center) to (23.center);
		\draw [in=-90, out=0] (22.center) to (23.center);
		\draw (1) to (21.center);
		\draw (20) to (22.center);
		\draw (11.center) to (25.center);
		\draw (15.center) to (24.center);
	\end{pgfonlayer}
\end{tikzpicture}

}

\newcommand{\conceptAlignmentMLE}{
\begin{tikzpicture}
	\begin{pgfonlayer}{nodelayer}
		\node [style=small box] (1) at (4.175, 3.25) {$\pdist_{\rv{C} \mid \text{pa}(\rv{C}); \Theta}$};
		\node [style=none] (6) at (0, 3.125) {$\text{pa}(\rv{C})$};
		\node [style=black dot] (7) at (1.25, 3.825) {};
		\node [style=none] (10) at (2.55, 4.275) {};
		\node [style=none] (11) at (2.55, 3.375) {};
		\node [style=white dot] (12) at (2.25, 3.825) {};
		\node [style=triangleflip] (14) at (6.6, 4.275) {$\hat{\theta}$};
		\node [style=none] (15) at (0.5, 3.125) {};
		\node [style=triangle] (20) at (0.1, 2.25) {$c^*$};
		\node [style=none] (21) at (5.1, 3.25) {};
		\node [style=none] (22) at (5.1, 2.25) {};
		\node [style=none] (23) at (5.675, 2.75) {};
		\node [style=none] (24) at (3.2, 3.125) {};
		\node [style=none] (25) at (3.2, 3.375) {};
	\end{pgfonlayer}
	\begin{pgfonlayer}{edgelayer}
		\draw [in=90, out=180] (10.center) to (12);
		\draw [in=-90, out=-180] (11.center) to (12);
		\draw (7) to (12);
		\draw [in=90, out=0] (21.center) to (23.center);
		\draw [in=-90, out=0] (22.center) to (23.center);
		\draw (20) to (22.center);
		\draw (11.center) to (25.center);
		\draw (15.center) to (24.center);
		\draw (10.center) to (14);
		\draw (1) to (21.center);
	\end{pgfonlayer}
\end{tikzpicture}

}

\newcommand{\counterfactualModel}{
\begin{tikzpicture}
	\begin{pgfonlayer}{nodelayer}
		\node [style=none] (10) at (2.8, 4.275) {};
		\node [style=none] (11) at (2.8, 3.375) {};
		\node [style=white dot] (12) at (2.5, 3.825) {};
		\node [style=none] (14) at (5.025, 3.375) {};
		\node [style=small box] (15) at (1.5, 3.825) {$\pdist_{\rv{C}_1 \mid \rv{U_1}}$};
		\node [style=small box] (26) at (0.1, 3.825) {$\pdist_{\rv{U}_1}$};
		\node [style=none] (28) at (2.975, 4.525) {};
		\node [style=small box] (29) at (0.1, 4.525) {$\pdist_{\rv{U}_2}$};
		\node [style=none] (31) at (5.275, 3.375) {$\rv{C}_1$};
		\node [style=small box] (32) at (3.85, 4.4) {$\pdist_{\rv{C}_2 \mid \rv{C}_1, \rv{U_2}}$};
		\node [style=none] (33) at (5.025, 4.4) {};
		\node [style=none] (34) at (5.275, 4.4) {$\rv{C}_2$};
		\node [style=none] (35) at (2.975, 4.275) {};
	\end{pgfonlayer}
	\begin{pgfonlayer}{edgelayer}
		\draw [in=90, out=180] (10.center) to (12);
		\draw [in=-90, out=-180] (11.center) to (12);
		\draw (15) to (12);
		\draw (26) to (15);
		\draw (29) to (28.center);
		\draw (32) to (33.center);
		\draw (10.center) to (35.center);
		\draw (11.center) to (14.center);
	\end{pgfonlayer}
\end{tikzpicture}

}

\newcommand{\counterfactualInference}{
\begin{tikzpicture}
	\begin{pgfonlayer}{nodelayer}
		\node [style=none] (10) at (2.8, 4.275) {};
		\node [style=none] (11) at (2.8, 3.375) {};
		\node [style=white dot] (12) at (2.5, 3.825) {};
		\node [style=small box] (15) at (1.5, 3.825) {$\pdist_{\rv{C}_1 \mid \rv{U_1}}$};
		\node [style=small box] (26) at (-0.1, 4.275) {$\pdist_{\rv{U}_1}$};
		\node [style=none] (28) at (2.975, 4.525) {};
		\node [style=small box] (29) at (-0.1, 5.425) {$\pdist_{\rv{U}_2}$};
		\node [style=small box] (32) at (3.85, 4.4) {$\pdist_{\rv{C}_2 \mid \rv{C}_1, \rv{U_2}}$};
		\node [style=none] (35) at (2.975, 4.275) {};
		\node [style=none] (36) at (5.025, 3.375) {};
		\node [style=none] (37) at (5.025, 2.575) {};
		\node [style=none] (38) at (5.5, 2.975) {};
		\node [style=none] (39) at (5.025, 4.4) {};
		\node [style=none] (40) at (5.025, 1.6) {};
		\node [style=none] (41) at (6, 3) {};
		\node [style=none] (42) at (0.85, 5.875) {};
		\node [style=none] (43) at (0.85, 4.975) {};
		\node [style=white dot] (44) at (0.55, 5.425) {};
		\node [style=none] (45) at (0.85, 4.725) {};
		\node [style=none] (46) at (0.85, 3.825) {};
		\node [style=white dot] (47) at (0.55, 4.275) {};
		\node [style=none] (48) at (5.65, 0.575) {};
		\node [style=small box] (52) at (6, 5.625) {$\pdist_{\rv{C}_1^* \mid \rv{U_1}}$};
		\node [style=none] (53) at (7.95, 6.375) {};
		\node [style=small box] (55) at (8.85, 6.25) {$\pdist_{\rv{C}_2^* \mid \rv{C}_1^*, \rv{U_2}}$};
		\node [style=none] (56) at (10.025, 6.25) {};
		\node [style=none] (57) at (10.275, 6.25) {$\rv{C}_2^*$};
		\node [style=none] (58) at (7.95, 6.125) {};
		\node [style=triangle] (59) at (-1.35, 2.575) {$k_1$};
		\node [style=triangle] (60) at (-1.35, 1.575) {$k_2$};
		\node [style=none] (61) at (6.775, 5.625) {};
		\node [style=none] (62) at (6.775, 5.625) {};
		\node [style=none] (63) at (6.775, 5.375) {};
		\node [style=none] (64) at (6.775, 5.875) {};
		\node [style=none] (65) at (6.875, 5.475) {};
		\node [style=none] (66) at (6.875, 5.775) {};
		\node [style=none] (67) at (6.975, 5.575) {};
		\node [style=none] (68) at (6.975, 5.675) {};
		\node [style=triangle] (69) at (-1.35, 0.575) {$k_1^*$};
		\node [style=none] (70) at (2.8, 6.375) {};
		\node [style=none] (71) at (-0.725, 6.75) {};
		\node [style=none] (72) at (-0.725, 0.325) {};
		\node [style=none] (73) at (9.9, 6.75) {};
		\node [style=none] (74) at (9.9, 0.325) {};
	\end{pgfonlayer}
	\begin{pgfonlayer}{edgelayer}
		\draw [in=90, out=180] (10.center) to (12);
		\draw [in=-90, out=-180] (11.center) to (12);
		\draw (15) to (12);
		\draw (10.center) to (35.center);
		\draw [in=90, out=0] (36.center) to (38.center);
		\draw [in=-90, out=0] (37.center) to (38.center);
		\draw [in=90, out=0] (39.center) to (41.center);
		\draw [in=-90, out=0] (40.center) to (41.center);
		\draw [in=90, out=180] (42.center) to (44);
		\draw [in=-90, out=-180] (43.center) to (44);
		\draw [in=90, out=180] (45.center) to (47);
		\draw [in=-90, out=-180] (46.center) to (47);
		\draw (29) to (44);
		\draw (26) to (47);
		\draw (46.center) to (15);
		\draw (55) to (56.center);
		\draw [in=180, out=0, looseness=0.50] (48.center) to (58.center);
		\draw [in=180, out=0] (43.center) to (28.center);
		\draw [in=-180, out=0] (45.center) to (52);
		\draw (59) to (37.center);
		\draw (11.center) to (36.center);
		\draw (32) to (39.center);
		\draw (60) to (40.center);
		\draw [style=thick edge] (64.center) to (63.center);
		\draw [style=thick edge] (66.center) to (65.center);
		\draw [style=thick edge] (68.center) to (67.center);
		\draw (52) to (62.center);
		\draw (69) to (48.center);
		\draw [in=180, out=0] (42.center) to (70.center);
		\draw (70.center) to (53.center);
		\draw [style=norm] (71.center) to (72.center);
		\draw [style=norm] (73.center) to (74.center);
		\draw [style=norm] (72.center) to (74.center);
		\draw [style=norm] (71.center) to (73.center);
	\end{pgfonlayer}
\end{tikzpicture}

}

\newcommand{\counterfactualInferenceAbduction}{
\begin{tikzpicture}
	\begin{pgfonlayer}{nodelayer}
		\node [style=small box] (0) at (6.1, 3.375) {$\pdist_{\rv{Y} \mid \rv{C}}$};
		\node [style=small box] (2) at (2, 3.725) {$\pdist_{\rv{U}}$};
		\node [style=none] (3) at (3.3, 4.075) {};
		\node [style=none] (4) at (3.3, 3.375) {};
		\node [style=white dot] (5) at (3, 3.725) {};
		\node [style=none] (6) at (7.625, 4.075) {};
		\node [style=none] (7) at (7.875, 4.075) {$\s{U}$};
		\node [style=none] (9) at (1.5, 4.3) {};
		\node [style=none] (10) at (1.5, 2.325) {};
		\node [style=none] (11) at (7.375, 4.3) {};
		\node [style=none] (12) at (7.375, 2.325) {};
		\node [style=triangle] (13) at (0.85, 2.575) {$y$};
		\node [style=none] (14) at (6.625, 3.375) {};
		\node [style=none] (15) at (6.625, 2.575) {};
		\node [style=none] (16) at (7.2, 2.975) {};
		\node [style=small box] (17) at (4.4, 3.375) {$\pdist_{\rv{C}_1,\dots,\rv{C}_n \mid \rv{U}}$};
	\end{pgfonlayer}
	\begin{pgfonlayer}{edgelayer}
		\draw [in=90, out=180] (3.center) to (5);
		\draw [in=-90, out=-180] (4.center) to (5);
		\draw (2) to (5);
		\draw (3.center) to (6.center);
		\draw [style=norm] (9.center) to (10.center);
		\draw [style=norm] (11.center) to (12.center);
		\draw [style=norm] (10.center) to (12.center);
		\draw [style=norm] (9.center) to (11.center);
		\draw [in=90, out=0] (14.center) to (16.center);
		\draw [in=-90, out=0] (15.center) to (16.center);
		\draw (0) to (14.center);
		\draw (13) to (15.center);
		\draw (4.center) to (17);
		\draw (17) to (0);
	\end{pgfonlayer}
\end{tikzpicture}

}

\newcommand{\counterfactualInferenceAction}{
\begin{tikzpicture}
	\begin{pgfonlayer}{nodelayer}
		\node [style=none] (0) at (0.6, 3.825) {$\s{U}$};
		\node [style=black dot] (1) at (2.6, 3.825) {};
		\node [style=none] (2) at (3.65, 4.175) {};
		\node [style=none] (3) at (3.65, 3.475) {};
		\node [style=white dot] (4) at (3.35, 3.825) {};
		\node [style=none] (5) at (6.25, 4.175) {};
		\node [style=none] (6) at (6.5, 4.175) {$\s{C}_i$};
		\node [style=none] (7) at (1.1, 3.825) {};
		\node [style=none] (8) at (1.25, 4.4) {};
		\node [style=none] (9) at (1.25, 2.625) {};
		\node [style=none] (10) at (6, 4.4) {};
		\node [style=none] (11) at (6, 2.625) {};
		\node [style=none] (12) at (5.35, 3.475) {};
		\node [style=none] (13) at (5.35, 2.875) {};
		\node [style=none] (14) at (5.825, 3.175) {};
		\node [style=triangle] (15) at (0.5, 2.875) {$k$};
		\node [style=none] (16) at (1.75, 3.825) {};
		\node [style=none] (17) at (1.75, 3.825) {};
		\node [style=none] (18) at (1.75, 3.575) {};
		\node [style=none] (19) at (1.75, 4.075) {};
		\node [style=none] (20) at (1.85, 3.675) {};
		\node [style=none] (21) at (1.85, 3.975) {};
		\node [style=none] (22) at (1.95, 3.775) {};
		\node [style=none] (23) at (1.95, 3.875) {};
	\end{pgfonlayer}
	\begin{pgfonlayer}{edgelayer}
		\draw [in=90, out=180] (2.center) to (4);
		\draw [in=-90, out=-180] (3.center) to (4);
		\draw (1) to (4);
		\draw (2.center) to (5.center);
		\draw [style=norm] (8.center) to (9.center);
		\draw [style=norm] (10.center) to (11.center);
		\draw [style=norm] (9.center) to (11.center);
		\draw [style=norm] (8.center) to (10.center);
		\draw [in=90, out=0] (12.center) to (14.center);
		\draw [in=-90, out=0] (13.center) to (14.center);
		\draw (15) to (13.center);
		\draw [style=thick edge] (19.center) to (18.center);
		\draw [style=thick edge] (21.center) to (20.center);
		\draw [style=thick edge] (23.center) to (22.center);
		\draw (7.center) to (17.center);
		\draw (3.center) to (12.center);
	\end{pgfonlayer}
\end{tikzpicture}

}

\newcommand{\pointwiseProbabilisticInference}{
\begin{tikzpicture}
	\begin{pgfonlayer}{nodelayer}
		\node [style=triangle] (5) at (7.475, 2.625) {$x$};
		\node [style=small box] (10) at (9.725, 3.25) {$\pdist_{\rv{Y} \mid \rv{X}; \Theta}$};
		\node [style=none] (11) at (10.975, 3.25) {};
		\node [style=none] (12) at (11.275, 3.25) {$\rv{Y}$};
		\node [style=white dot] (14) at (8.525, 2.625) {};
		\node [style=none] (17) at (10.975, 2.625) {};
		\node [style=none] (18) at (10.975, 2.625) {};
		\node [style=none] (19) at (10.975, 2.625) {};
		\node [style=none] (20) at (10.975, 2.375) {};
		\node [style=none] (21) at (10.975, 2.875) {};
		\node [style=none] (22) at (11.075, 2.475) {};
		\node [style=none] (23) at (11.075, 2.775) {};
		\node [style=none] (24) at (11.175, 2.575) {};
		\node [style=none] (25) at (11.175, 2.675) {};
	\end{pgfonlayer}
	\begin{pgfonlayer}{edgelayer}
		\draw (10) to (11.center);
		\draw (5) to (14);
		\draw [in=-165, out=15] (14) to (10);
		\draw (14) to (17.center);
		\draw [style=thick edge] (21.center) to (20.center);
		\draw [style=thick edge] (23.center) to (22.center);
		\draw [style=thick edge] (25.center) to (24.center);
	\end{pgfonlayer}
\end{tikzpicture}

}

\newcommand{\inverseProbabilisticInference}{
\begin{tikzpicture}
	\begin{pgfonlayer}{nodelayer}
		\node [style=small box] (0) at (2.25, 3.25) {$\pdist_{\rv{Y} \mid \rv{X}}$};
		\node [style=none] (21) at (4.025, 3.75) {$X$};
		\node [style=none] (24) at (3.725, 3.75) {};a
		\node [style=triangle] (25) at (-0.75, 2.45) {$y$};
		\node [style=none] (29) at (3, 3.25) {};
		\node [style=none] (30) at (3, 2.45) {};
		\node [style=none] (31) at (3.4, 2.85) {};
		\node [style=small box] (32) at (0.475, 3.25) {$\pdist_{\rv{X}}$};
		\node [style=none] (33) at (-0.25, 4.25) {};
		\node [style=none] (34) at (-0.25, 2.25) {};
		\node [style=none] (35) at (3.525, 4.25) {};
		\node [style=none] (36) at (3.525, 2.25) {};
		\node [style=none] (37) at (1.575, 3.75) {};
		\node [style=white dot] (38) at (1.275, 3.25) {};
	\end{pgfonlayer}
	\begin{pgfonlayer}{edgelayer}
		\draw [in=90, out=0] (29.center) to (31.center);
		\draw [in=-90, out=0] (30.center) to (31.center);
		\draw (0) to (29.center);
		\draw (25) to (30.center);
		\draw [style=norm] (33.center) to (35.center);
		\draw [style=norm] (35.center) to (36.center);
		\draw [style=norm] (33.center) to (34.center);
		\draw [style=norm] (34.center) to (36.center);
		\draw [in=90, out=180] (37.center) to (38);
		\draw (37.center) to (24.center);
		\draw (32) to (38);
		\draw (38) to (0);
	\end{pgfonlayer}
\end{tikzpicture}

}

\newcommand{\exampleConceptRound}{
\begin{tikzpicture}[
    node distance = .3cm,
]

\node[circle, draw, thick, fill=gray!30, fill opacity=0.3, minimum size=3cm] (models) {};

\node[circle, draw, thick, fill=gray!30, fill opacity=0.3, minimum size=3cm, right=1cm of models] (sentences) {};

\node (apple) at (-.3,.3) {\inlineimg{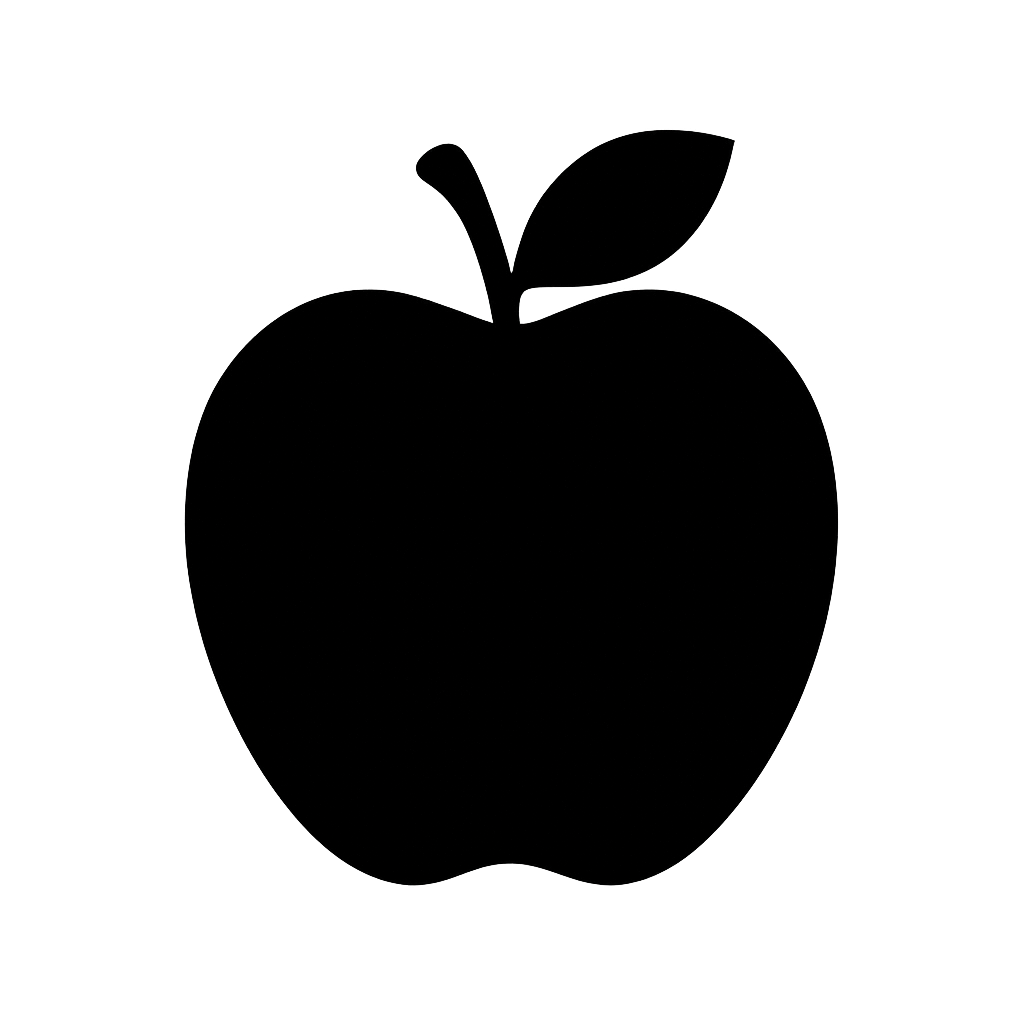}};
\node (donut) at (.3,.5) {\inlineimg{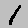}};
\node (ball) at (.4,-.1) {\inlineimg{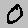}};
\node (avocado) at (-.3,-.9) {\inlineimg{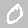}};

\node (green) at (4.7,-.1) {\text{white}};
\node (round) at (3.9,.8) {\text{black}};
\node (sweet) at (3.3,-.2) {\text{round}};
\node (soft) at (4.4,-.7) {\text{even}};

\node (U) at (-1.6,.8) {$\mathcal{U}$};
\node (S) at (5.6,.8) {$\mathcal{S}$};

\begin{pgfonlayer}{background}
    \node[draw, thick, fill=gray!30, rounded corners, inner sep=.2cm, yshift=.1cm, fit=(apple) (donut) (ball), label=above:{$\s{M}$}] (M) {};

    \node[draw, thick, fill=gray!30, rounded corners, inner sep=.cm, yshift=.cm, fit=(round), label=right:{$\s{T}$}] (T) {};

    \draw[arrowstyledash, blue] (T)  to [bend left=20] node[midway, below] {$\beta$} (M);
    \draw[arrowstyledash, red] (M)  to [bend left=20] node[midway, above] {$\gamma$} (T);
\end{pgfonlayer}

\end{tikzpicture}
}

\newcommand{\interventionModel}{
\begin{tikzpicture}
	\begin{pgfonlayer}{nodelayer}
		\node [style=none] (171) at (13.825, 3.3) {$\rv{V}_{i,\rv{T}}$};
		\node [style=none] (174) at (9.825, 3.3) {};
		\node [style=none] (175) at (9.375, 3.3) {$\rv{V}_{k}$};
		\node [style=none] (176) at (13.5, 3.3) {};
		\node [style=small box] (187) at (11.55, 3.3) {$\pdist_{\rv{V}_{i, \rv{T}} \mid \rv{V}_k}$};
	\end{pgfonlayer}
	\begin{pgfonlayer}{edgelayer}
		\draw (174.center) to (187);
		\draw (187) to (176.center);
	\end{pgfonlayer}
\end{tikzpicture}

}

\newcommand{\localIntervention}{
\begin{tikzpicture}
	\begin{pgfonlayer}{nodelayer}
		\node [style=none] (168) at (9.575, 1.95) {};
		\node [style=none] (169) at (9.125, 1.95) {$\rv{V}_{j}$};
		\node [style=none] (171) at (15.575, 4) {$\rv{V}_{i,\rv{T}}$};
		\node [style=small box] (172) at (13.45, 2.75) {$\pdist_{\rv{V}_{j} \mid \rv{V}_{i,\rv{T}}}$};
		\node [style=none] (174) at (9.575, 3) {};
		\node [style=none] (175) at (9.125, 3) {$\rv{V}_{k}$};
		\node [style=none] (176) at (15.25, 4) {};
		\node [style=none] (177) at (10.075, 4.25) {};
		\node [style=none] (178) at (14.825, 4.25) {};
		\node [style=none] (179) at (10.075, 1.75) {};
		\node [style=none] (180) at (14.825, 1.75) {};
		\node [style=none] (181) at (10.675, 4) {};
		\node [style=none] (182) at (10.675, 3.3) {};
		\node [style=none] (183) at (10.325, 3.65) {};
		\node [style=none] (184) at (14.325, 2.75) {};
		\node [style=none] (185) at (14.325, 1.95) {};
		\node [style=none] (186) at (14.725, 2.35) {};
		\node [style=small box] (187) at (11.55, 3.3) {$\pdist_{\rv{V}_{i, \rv{T}} \mid \rv{V}_k}$};
	\end{pgfonlayer}
	\begin{pgfonlayer}{edgelayer}
		\draw [style=norm] (177.center) to (179.center);
		\draw [style=norm] (179.center) to (180.center);
		\draw [style=norm] (180.center) to (178.center);
		\draw [style=norm] (178.center) to (177.center);
		\draw [in=90, out=180] (181.center) to (183.center);
		\draw [in=-90, out=-180] (182.center) to (183.center);
		\draw [in=90, out=0] (184.center) to (186.center);
		\draw [in=-90, out=0] (185.center) to (186.center);
		\draw (182.center) to (187);
		\draw (181.center) to (176.center);
		\draw [in=-165, out=0, looseness=1.25] (174.center) to (187);
		\draw (172) to (184.center);
		\draw (168.center) to (185.center);
		\draw [in=180, out=0] (187) to (172);
	\end{pgfonlayer}
\end{tikzpicture}

}

\newcommand{\breakIntervention}{
\begin{tikzpicture}
	\begin{pgfonlayer}{nodelayer}
		\node [style=none] (0) at (12.825, 4) {$\rv{V}_{i,\rv{T}}$};
		\node [style=none] (1) at (9.575, 3) {};
		\node [style=none] (2) at (9.125, 3) {$\rv{V}_{k}$};
		\node [style=none] (3) at (12.5, 4) {};
		\node [style=small box] (7) at (10.675, 4) {$\pdist_{\rv{V}_{i,\rv{T}}}$};
		\node [style=none] (8) at (10.575, 3) {};
		\node [style=none] (9) at (10.575, 3) {};
		\node [style=none] (10) at (10.575, 2.75) {};
		\node [style=none] (11) at (10.575, 3.25) {};
		\node [style=none] (12) at (10.675, 2.85) {};
		\node [style=none] (13) at (10.675, 3.15) {};
		\node [style=none] (14) at (10.775, 2.95) {};
		\node [style=none] (15) at (10.775, 3.05) {};
	\end{pgfonlayer}
	\begin{pgfonlayer}{edgelayer}
		\draw [style=thick edge] (11.center) to (10.center);
		\draw [style=thick edge] (13.center) to (12.center);
		\draw [style=thick edge] (15.center) to (14.center);
		\draw (1.center) to (9.center);
		\draw (7) to (3.center);
	\end{pgfonlayer}
\end{tikzpicture}

}

\newcommand{\doIntervention}{
\begin{tikzpicture}
	\begin{pgfonlayer}{nodelayer}
		\node [style=none] (1) at (0.6, 3.825) {$\text{pa}(\s{C})$};
		\node [style=black dot] (2) at (2.6, 3.825) {};
		\node [style=none] (3) at (3.65, 4.275) {};
		\node [style=none] (4) at (3.65, 3.375) {};
		\node [style=white dot] (5) at (3.35, 3.825) {};
		\node [style=none] (6) at (6.25, 4.275) {};
		\node [style=none] (7) at (6.5, 4.275) {$\s{C}$};
		\node [style=none] (8) at (1.1, 3.825) {};
		\node [style=none] (9) at (1.25, 4.5) {};
		\node [style=none] (10) at (1.25, 2.325) {};
		\node [style=none] (11) at (6, 4.5) {};
		\node [style=none] (12) at (6, 2.325) {};
		\node [style=none] (13) at (5.35, 3.375) {};
		\node [style=none] (14) at (5.35, 2.575) {};
		\node [style=none] (15) at (5.825, 2.975) {};
		\node [style=triangle] (17) at (0.5, 2.575) {$k$};
		\node [style=none] (18) at (1.75, 3.825) {};
		\node [style=none] (19) at (1.75, 3.825) {};
		\node [style=none] (20) at (1.75, 3.575) {};
		\node [style=none] (21) at (1.75, 4.075) {};
		\node [style=none] (22) at (1.85, 3.675) {};
		\node [style=none] (23) at (1.85, 3.975) {};
		\node [style=none] (24) at (1.95, 3.775) {};
		\node [style=none] (25) at (1.95, 3.875) {};
	\end{pgfonlayer}
	\begin{pgfonlayer}{edgelayer}
		\draw [in=90, out=180] (3.center) to (5);
		\draw [in=-90, out=-180] (4.center) to (5);
		\draw (2) to (5);
		\draw (3.center) to (6.center);
		\draw [style=norm] (9.center) to (10.center);
		\draw [style=norm] (11.center) to (12.center);
		\draw [style=norm] (10.center) to (12.center);
		\draw [style=norm] (9.center) to (11.center);
		\draw [in=90, out=0] (13.center) to (15.center);
		\draw [in=-90, out=0] (14.center) to (15.center);
		\draw (17) to (14.center);
		\draw [style=thick edge] (21.center) to (20.center);
		\draw [style=thick edge] (23.center) to (22.center);
		\draw [style=thick edge] (25.center) to (24.center);
		\draw (8.center) to (19.center);
		\draw (4.center) to (13.center);
	\end{pgfonlayer}
\end{tikzpicture}

}

\newcommand{\causalEffectEstimation}{
\begin{tikzpicture}
	\begin{pgfonlayer}{nodelayer}
		\node [style=none] (1) at (0.6, 3.825) {$\text{pa}(\rv{C})$};
		\node [style=black dot] (2) at (2.6, 3.825) {};
		\node [style=none] (3) at (3.65, 4.275) {};
		\node [style=none] (4) at (3.65, 3.375) {};
		\node [style=white dot] (5) at (3.35, 3.825) {};
		\node [style=none] (6) at (6.25, 4.275) {};
		\node [style=none] (7) at (6.75, 4.275) {$\rv{C}_{[k=0]}$};
		\node [style=none] (8) at (1.1, 3.825) {};
		\node [style=none] (9) at (1.25, 4.5) {};
		\node [style=none] (10) at (1.25, 2.325) {};
		\node [style=none] (11) at (6, 4.5) {};
		\node [style=none] (12) at (6, 2.325) {};
		\node [style=none] (13) at (5.35, 3.375) {};
		\node [style=none] (14) at (5.35, 2.575) {};
		\node [style=none] (15) at (5.825, 2.975) {};
		\node [style=triangle] (17) at (0.5, 2.575) {$0$};
		\node [style=none] (18) at (1.75, 3.825) {};
		\node [style=none] (19) at (1.75, 3.825) {};
		\node [style=none] (20) at (1.75, 3.575) {};
		\node [style=none] (21) at (1.75, 4.075) {};
		\node [style=none] (22) at (1.85, 3.675) {};
		\node [style=none] (23) at (1.85, 3.975) {};
		\node [style=none] (24) at (1.95, 3.775) {};
		\node [style=none] (25) at (1.95, 3.875) {};
		\node [style=none] (26) at (0.6, 0.575) {$\text{pa}(\rv{C})$};
		\node [style=black dot] (27) at (2.6, 0.575) {};
		\node [style=none] (28) at (3.65, 1.025) {};
		\node [style=none] (29) at (3.65, 0.125) {};
		\node [style=white dot] (30) at (3.35, 0.575) {};
		\node [style=none] (31) at (6.25, 1.025) {};
		\node [style=none] (32) at (6.75, 1.025) {$\rv{C}_{[k=1]}$};
		\node [style=none] (33) at (1.1, 0.575) {};
		\node [style=none] (34) at (1.25, 1.25) {};
		\node [style=none] (35) at (1.25, -0.925) {};
		\node [style=none] (36) at (6, 1.25) {};
		\node [style=none] (37) at (6, -0.925) {};
		\node [style=none] (38) at (5.35, 0.125) {};
		\node [style=none] (39) at (5.35, -0.675) {};
		\node [style=none] (40) at (5.825, -0.275) {};
		\node [style=triangle] (41) at (0.5, -0.675) {$1$};
		\node [style=none] (42) at (1.75, 0.575) {};
		\node [style=none] (43) at (1.75, 0.575) {};
		\node [style=none] (44) at (1.75, 0.325) {};
		\node [style=none] (45) at (1.75, 0.825) {};
		\node [style=none] (46) at (1.85, 0.425) {};
		\node [style=none] (47) at (1.85, 0.725) {};
		\node [style=none] (48) at (1.95, 0.525) {};
		\node [style=none] (49) at (1.95, 0.625) {};
	\end{pgfonlayer}
	\begin{pgfonlayer}{edgelayer}
		\draw [in=90, out=180] (3.center) to (5);
		\draw [in=-90, out=-180] (4.center) to (5);
		\draw (2) to (5);
		\draw (3.center) to (6.center);
		\draw [style=norm] (9.center) to (10.center);
		\draw [style=norm] (11.center) to (12.center);
		\draw [style=norm] (10.center) to (12.center);
		\draw [style=norm] (9.center) to (11.center);
		\draw [in=90, out=0] (13.center) to (15.center);
		\draw [in=-90, out=0] (14.center) to (15.center);
		\draw (17) to (14.center);
		\draw [style=thick edge] (21.center) to (20.center);
		\draw [style=thick edge] (23.center) to (22.center);
		\draw [style=thick edge] (25.center) to (24.center);
		\draw (8.center) to (19.center);
		\draw (4.center) to (13.center);
		\draw [in=90, out=180] (28.center) to (30);
		\draw [in=-90, out=-180] (29.center) to (30);
		\draw (27) to (30);
		\draw (28.center) to (31.center);
		\draw [style=norm] (34.center) to (35.center);
		\draw [style=norm] (36.center) to (37.center);
		\draw [style=norm] (35.center) to (37.center);
		\draw [style=norm] (34.center) to (36.center);
		\draw [in=90, out=0] (38.center) to (40.center);
		\draw [in=-90, out=0] (39.center) to (40.center);
		\draw (41) to (39.center);
		\draw [style=thick edge] (45.center) to (44.center);
		\draw [style=thick edge] (47.center) to (46.center);
		\draw [style=thick edge] (49.center) to (48.center);
		\draw (33.center) to (43.center);
		\draw (29.center) to (38.center);
	\end{pgfonlayer}
\end{tikzpicture}

}

\newcommand{\replaceIntervention}{
\begin{tikzpicture}
	\begin{pgfonlayer}{nodelayer}
		\node [style=none] (168) at (9.575, 1.95) {};
		\node [style=none] (169) at (9.125, 1.95) {$\rv{V}_{j}$};
		\node [style=none] (171) at (15.575, 4) {$\rv{V}_{i,\rv{T}}$};
		\node [style=small box] (172) at (13.45, 2.75) {$\pdist_{\rv{V}_{j} \mid \rv{V}_{i,\rv{T}}}$};
		\node [style=none] (174) at (9.575, 2.75) {};
		\node [style=none] (175) at (9.125, 2.75) {$\rv{V}_{k}$};
		\node [style=none] (176) at (15.25, 4) {};
		\node [style=none] (177) at (10.075, 4.25) {};
		\node [style=none] (178) at (14.825, 4.25) {};
		\node [style=none] (179) at (10.075, 1.75) {};
		\node [style=none] (180) at (14.825, 1.75) {};
		\node [style=none] (181) at (10.675, 4) {};
		\node [style=none] (182) at (10.675, 3.3) {};
		\node [style=none] (183) at (10.325, 3.65) {};
		\node [style=none] (184) at (14.325, 2.75) {};
		\node [style=none] (185) at (14.325, 1.95) {};
		\node [style=none] (186) at (14.725, 2.35) {};
		\node [style=small box] (187) at (11.55, 3.3) {$\pdist_{\rv{V}_{i, \rv{T}} \mid \rv{V}_k}$};
		\node [style=none] (188) at (10.575, 2.75) {};
		\node [style=none] (189) at (10.575, 2.75) {};
		\node [style=none] (190) at (10.575, 2.5) {};
		\node [style=none] (191) at (10.575, 3) {};
		\node [style=none] (192) at (10.675, 2.6) {};
		\node [style=none] (193) at (10.675, 2.9) {};
		\node [style=none] (194) at (10.775, 2.7) {};
		\node [style=none] (195) at (10.775, 2.8) {};
	\end{pgfonlayer}
	\begin{pgfonlayer}{edgelayer}
		\draw [style=norm] (177.center) to (179.center);
		\draw [style=norm] (179.center) to (180.center);
		\draw [style=norm] (180.center) to (178.center);
		\draw [style=norm] (178.center) to (177.center);
		\draw [in=90, out=180] (181.center) to (183.center);
		\draw [in=-90, out=-180] (182.center) to (183.center);
		\draw [in=90, out=0] (184.center) to (186.center);
		\draw [in=-90, out=0] (185.center) to (186.center);
		\draw (182.center) to (187);
		\draw (181.center) to (176.center);
		\draw (172) to (184.center);
		\draw (168.center) to (185.center);
		\draw [in=180, out=0] (187) to (172);
		\draw [style=thick edge] (191.center) to (190.center);
		\draw [style=thick edge] (193.center) to (192.center);
		\draw [style=thick edge] (195.center) to (194.center);
		\draw (174.center) to (189.center);
	\end{pgfonlayer}
\end{tikzpicture}

}

\newcommand{\groundTruthIntervention}{
\begin{tikzpicture}
	\begin{pgfonlayer}{nodelayer}
		\node [style=none] (1) at (0.6, 3.825) {$\text{pa}(\s{C})$};
		\node [style=black dot] (2) at (2.6, 3.825) {};
		\node [style=none] (3) at (3.65, 4.275) {};
		\node [style=none] (4) at (3.65, 3.375) {};
		\node [style=white dot] (5) at (3.35, 3.825) {};
		\node [style=none] (6) at (6.25, 4.275) {};
		\node [style=none] (7) at (6.5, 4.275) {$\s{C}$};
		\node [style=none] (8) at (1.1, 3.825) {};
		\node [style=none] (9) at (1.25, 4.5) {};
		\node [style=none] (10) at (1.25, 2.325) {};
		\node [style=none] (11) at (6, 4.5) {};
		\node [style=none] (12) at (6, 2.325) {};
		\node [style=none] (13) at (5.35, 3.375) {};
		\node [style=none] (14) at (5.35, 2.575) {};
		\node [style=none] (15) at (5.825, 2.975) {};
		\node [style=triangle] (17) at (0.5, 2.575) {$c^*$};
		\node [style=none] (18) at (1.75, 3.825) {};
		\node [style=none] (19) at (1.75, 3.825) {};
		\node [style=none] (20) at (1.75, 3.575) {};
		\node [style=none] (21) at (1.75, 4.075) {};
		\node [style=none] (22) at (1.85, 3.675) {};
		\node [style=none] (23) at (1.85, 3.975) {};
		\node [style=none] (24) at (1.95, 3.775) {};
		\node [style=none] (25) at (1.95, 3.875) {};
	\end{pgfonlayer}
	\begin{pgfonlayer}{edgelayer}
		\draw [in=90, out=180] (3.center) to (5);
		\draw [in=-90, out=-180] (4.center) to (5);
		\draw (2) to (5);
		\draw (3.center) to (6.center);
		\draw [style=norm] (9.center) to (10.center);
		\draw [style=norm] (11.center) to (12.center);
		\draw [style=norm] (10.center) to (12.center);
		\draw [style=norm] (9.center) to (11.center);
		\draw [in=90, out=0] (13.center) to (15.center);
		\draw [in=-90, out=0] (14.center) to (15.center);
		\draw (17) to (14.center);
		\draw [style=thick edge] (21.center) to (20.center);
		\draw [style=thick edge] (23.center) to (22.center);
		\draw [style=thick edge] (25.center) to (24.center);
		\draw (8.center) to (19.center);
		\draw (4.center) to (13.center);
	\end{pgfonlayer}
\end{tikzpicture}

}

\newcommand{\compositionalElements}{
\begin{tikzpicture}
	\begin{pgfonlayer}{nodelayer}
		\node [style=none] (17) at (-1.5, 7) {};
		\node [style=none] (21) at (1.25, 7) {};
		\node [style=none] (39) at (-3.75, 6) {Concept-based model};
		\node [style=none] (40) at (-0.75, 6) {Copy Map};
		\node [style=none] (41) at (2, 6) {Discard Map};
		\node [style=none] (43) at (2.325, 7) {};
		\node [style=none] (44) at (2.325, 7) {};
		\node [style=none] (45) at (2.325, 6.75) {};
		\node [style=none] (46) at (2.325, 7.25) {};
		\node [style=none] (47) at (2.425, 6.85) {};
		\node [style=none] (48) at (2.425, 7.15) {};
		\node [style=none] (49) at (2.525, 6.95) {};
		\node [style=none] (50) at (2.525, 7.05) {};
		\node [style=none] (51) at (-0.275, 7.5) {};
		\node [style=none] (52) at (-0.275, 6.5) {};
		\node [style=white dot] (53) at (-0.575, 7) {};
		\node [style=small box] (54) at (-3.75, 7) {$\pdist_{\rv{C}_i \mid \rv{C}_j}$};
		\node [style=none] (58) at (-2.575, 7) {$\s{C}_i$};
		\node [style=none] (59) at (-2.825, 7) {};
		\node [style=none] (61) at (-4.625, 7) {};
		\node [style=none] (68) at (-4.875, 7) {$\s{C}_j$};
	\end{pgfonlayer}
	\begin{pgfonlayer}{edgelayer}
		\draw [style=thick edge] (46.center) to (45.center);
		\draw [style=thick edge] (48.center) to (47.center);
		\draw [style=thick edge] (50.center) to (49.center);
		\draw (21.center) to (44.center);
		\draw [in=90, out=180] (51.center) to (53);
		\draw [in=-90, out=-180] (52.center) to (53);
		\draw (17.center) to (53);
		\draw (54) to (59.center);
		\draw (61.center) to (54);
	\end{pgfonlayer}
\end{tikzpicture}

}

\newcommand{\blueprint}{
\begin{tikzpicture}
	\begin{pgfonlayer}{nodelayer}
		\node [style=small box] (30) at (10.95, 3.35) {$\pdist_{\rv{C} \mid \rv{U}}$};
		\node [style=small box] (38) at (12.9, 3.875) {$\pdist_{\rv{C'} \mid \rv{C}, \rv{U}}$};
		\node [style=small box] (39) at (7.7, 3.95) {$\pdist_{\rv{Z} \mid \rv{X}}$};
		\node [style=small box] (40) at (15.225, 4.4) {$\pdist_{\rv{Y} \mid \rv{C}',\rv{U}}$};
		\node [style=none] (42) at (16.475, 4.4) {};
		\node [style=none] (43) at (16.725, 4.4) {$\rv{Y}$};
		\node [style=none] (44) at (6.325, 3.25) {};
		\node [style=none] (45) at (6.075, 3.25) {$\rv{X}$};
		\node [style=none] (46) at (7, 3.95) {};
		\node [style=none] (47) at (7, 2.425) {};
		\node [style=white dot] (48) at (6.7, 3.25) {};
		\node [style=none] (49) at (16.475, 2.4) {};
		\node [style=none] (50) at (16.725, 2.4) {$\rv{X}$};
		\node [style=none] (54) at (12.2, 3.8) {};
		\node [style=none] (55) at (12.175, 2.9) {};
		\node [style=white dot] (56) at (11.875, 3.35) {};
		\node [style=none] (57) at (14.55, 4.275) {};
		\node [style=none] (58) at (14.4, 3.425) {};
		\node [style=white dot] (59) at (14.1, 3.875) {};
		\node [style=none] (62) at (16.475, 2.9) {};
		\node [style=none] (63) at (16.725, 2.9) {$\rv{C}$};
		\node [style=none] (64) at (16.475, 3.425) {};
		\node [style=none] (65) at (16.725, 3.425) {$\rv{C'}$};
		\node [style=small box] (66) at (9, 3.95) {$\pdist_{\rv{U} \mid \rv{Z}}$};
		\node [style=none] (67) at (10.25, 4.525) {};
		\node [style=none] (68) at (10.25, 3.35) {};
		\node [style=white dot] (69) at (9.95, 3.95) {};
		\node [style=none] (70) at (14.55, 4.525) {};
		\node [style=none] (71) at (12.2, 3.95) {};
	\end{pgfonlayer}
	\begin{pgfonlayer}{edgelayer}
		\draw (40) to (42.center);
		\draw [in=90, out=-180] (46.center) to (48);
		\draw [in=-90, out=-180, looseness=0.75] (47.center) to (48);
		\draw (47.center) to (49.center);
		\draw (44.center) to (48);
		\draw (46.center) to (39);
		\draw [in=90, out=180] (54.center) to (56);
		\draw [in=-90, out=-180] (55.center) to (56);
		\draw [in=90, out=180] (57.center) to (59);
		\draw [in=-90, out=-180] (58.center) to (59);
		\draw (30) to (56);
		\draw (38) to (59);
		\draw (55.center) to (62.center);
		\draw (58.center) to (64.center);
		\draw (39) to (66);
		\draw [in=90, out=180] (67.center) to (69);
		\draw [in=-90, out=-180] (68.center) to (69);
		\draw (69) to (71.center);
		\draw (67.center) to (70.center);
		\draw (68.center) to (30);
		\draw (66) to (69);
	\end{pgfonlayer}
\end{tikzpicture}

}

\newcommand{\interventionPosterior}{
\begin{tikzpicture}
	\begin{pgfonlayer}{nodelayer}
		\node [style=small box] (0) at (3.925, 3.375) {$\pdist_{\text{pa}'(\rv{C}) \mid \rv{C}}$};
		\node [style=none] (1) at (0, 3.825) {$\text{pa}(\s{C})$};
		\node [style=small box] (2) at (1.75, 3.825) {$\pdist_{\rv{C} \mid \text{pa}(\rv{C})}$};
		\node [style=none] (3) at (3.05, 4.275) {};
		\node [style=none] (4) at (3.05, 3.375) {};
		\node [style=white dot] (5) at (2.75, 3.825) {};
		\node [style=none] (6) at (5.65, 4.275) {};
		\node [style=none] (7) at (5.9, 4.275) {$\s{C}$};
		\node [style=none] (8) at (0.5, 3.825) {};
		\node [style=none] (9) at (0.65, 4.5) {};
		\node [style=none] (10) at (0.65, 2.325) {};
		\node [style=none] (11) at (5.4, 4.5) {};
		\node [style=none] (12) at (5.4, 2.325) {};
		\node [style=none] (14) at (4.75, 3.375) {};
		\node [style=none] (15) at (4.75, 2.575) {};
		\node [style=none] (16) at (5.225, 2.975) {};
		\node [style=none] (19) at (-0.1, 2.575) {$\text{pa}'(\s{C})$};
		\node [style=none] (20) at (0.4, 2.575) {};
	\end{pgfonlayer}
	\begin{pgfonlayer}{edgelayer}
		\draw [in=90, out=180] (3.center) to (5);
		\draw [in=-90, out=-180] (4.center) to (5);
		\draw (2) to (5);
		\draw (3.center) to (6.center);
		\draw [style=norm] (9.center) to (10.center);
		\draw [style=norm] (11.center) to (12.center);
		\draw [style=norm] (10.center) to (12.center);
		\draw [style=norm] (9.center) to (11.center);
		\draw [in=90, out=0] (14.center) to (16.center);
		\draw [in=-90, out=0] (15.center) to (16.center);
		\draw (0) to (14.center);
		\draw (8.center) to (2);
		\draw (20.center) to (15.center);
		\draw (4.center) to (0);
	\end{pgfonlayer}
\end{tikzpicture}

}

\newcommand{\trainingIndependent}{
\begin{tikzpicture}
	\begin{pgfonlayer}{nodelayer}
		\node [style=small box] (0) at (3.675, 3.25) {$\pdist_{\rv{C}_\rv{T} \mid \rv{X}; \Theta_c}$};
		\node [style=triangle] (1) at (-0.025, 2.925) {$x$};
		\node [style=small box] (2) at (2.25, 3.7) {$\pdist_{\Theta_c}$};
		\node [style=triangle] (3) at (-0.025, 1.95) {$c$};
		\node [style=none] (4) at (4.5, 3.25) {};
		\node [style=none] (5) at (4.5, 2.45) {};
		\node [style=none] (6) at (4.9, 2.85) {};
		\node [style=none] (7) at (1.725, 4.65) {};
		\node [style=none] (8) at (1.725, 3.7) {};
		\node [style=none] (9) at (1.375, 4.175) {};
		\node [style=none] (10) at (1.225, 4.85) {};
		\node [style=none] (11) at (1.225, 1.7) {};
		\node [style=none] (14) at (10.4, 4.65) {};
		\node [style=none] (15) at (10.725, 4.65) {$\Theta_c$};
		\node [style=small box] (19) at (8.425, 3.25) {$\pdist_{\rv{Y}_\rv{T} \mid \rv{C}; \Theta_y}$};
		\node [style=small box] (21) at (7, 3.7) {$\pdist_{\Theta_y}$};
		\node [style=triangle] (22) at (-0.025, 0.95) {$y$};
		\node [style=none] (23) at (9.25, 3.25) {};
		\node [style=none] (24) at (9.25, 2.45) {};
		\node [style=none] (25) at (9.65, 2.85) {};
		\node [style=none] (26) at (6.475, 4.2) {};
		\node [style=none] (27) at (6.475, 3.7) {};
		\node [style=none] (28) at (6.125, 3.95) {};
		\node [style=none] (31) at (5.225, 4.85) {};
		\node [style=none] (32) at (5.225, 1.7) {};
		\node [style=none] (33) at (10.4, 4.2) {};
		\node [style=none] (34) at (10.725, 4.2) {$\Theta_y$};
		\node [style=none] (35) at (8.5, 2.45) {};
		\node [style=none] (36) at (6.475, 0.95) {};
		\node [style=none] (38) at (5.725, 1.45) {};
		\node [style=none] (39) at (5.975, 4.35) {};
		\node [style=none] (40) at (5.975, 0.7) {};
		\node [style=none] (41) at (9.975, 4.35) {};
		\node [style=none] (42) at (9.975, 0.7) {};
		\node [style=none] (43) at (1.025, 2.45) {};
		\node [style=none] (44) at (1.025, 1.45) {};
		\node [style=white dot] (45) at (0.725, 1.95) {};
	\end{pgfonlayer}
	\begin{pgfonlayer}{edgelayer}
		\draw [in=165, out=0, looseness=1.25] (2) to (0);
		\draw [in=-180, out=0, looseness=1.25] (1) to (0);
		\draw [in=90, out=0] (4.center) to (6.center);
		\draw [in=-90, out=0] (5.center) to (6.center);
		\draw (0) to (4.center);
		\draw [in=90, out=180] (7.center) to (9.center);
		\draw [in=-90, out=-180] (8.center) to (9.center);
		\draw (8.center) to (2);
		\draw [style=norm] (10.center) to (11.center);
		\draw (7.center) to (14.center);
		\draw [in=165, out=0, looseness=1.25] (21) to (19);
		\draw [in=90, out=0] (23.center) to (25.center);
		\draw [in=-90, out=0] (24.center) to (25.center);
		\draw (19) to (23.center);
		\draw [in=90, out=180] (26.center) to (28.center);
		\draw [in=-90, out=-180] (27.center) to (28.center);
		\draw (27.center) to (21);
		\draw [style=norm] (31.center) to (32.center);
		\draw (26.center) to (33.center);
		\draw (35.center) to (24.center);
		\draw [in=0, out=180, looseness=0.75] (35.center) to (36.center);
		\draw (22) to (36.center);
		\draw [style=norm] (11.center) to (32.center);
		\draw [style=norm] (10.center) to (31.center);
		\draw [in=0, out=-165, looseness=0.75] (19) to (38.center);
		\draw [style=norm] (39.center) to (40.center);
		\draw [style=norm] (41.center) to (42.center);
		\draw [style=norm] (40.center) to (42.center);
		\draw [style=norm] (39.center) to (41.center);
		\draw [in=90, out=180] (43.center) to (45);
		\draw [in=-90, out=-180] (44.center) to (45);
		\draw (44.center) to (38.center);
		\draw (43.center) to (5.center);
		\draw (3) to (45);
	\end{pgfonlayer}
\end{tikzpicture}

}

\newcommand{\trainingSequentialC}{
\begin{tikzpicture}
	\begin{pgfonlayer}{nodelayer}
		\node [style=small box] (0) at (3.675, 3.25) {$\pdist_{\rv{C}_\rv{T} \mid \rv{X}; \Theta_c}$};
		\node [style=triangle] (1) at (0.475, 2.925) {$x$};
		\node [style=small box] (2) at (2.25, 3.7) {$\pdist_{\Theta_c}$};
		\node [style=triangle] (3) at (0.475, 1.95) {$c$};
		\node [style=none] (4) at (4.5, 3.25) {};
		\node [style=none] (5) at (4.5, 1.95) {};
		\node [style=none] (6) at (4.9, 2.6) {};
		\node [style=none] (7) at (1.725, 4.65) {};
		\node [style=none] (8) at (1.725, 3.7) {};
		\node [style=none] (9) at (1.375, 4.175) {};
		\node [style=none] (10) at (1.225, 4.85) {};
		\node [style=none] (11) at (1.225, 1.7) {};
		\node [style=none] (14) at (5.9, 4.65) {};
		\node [style=none] (15) at (6.225, 4.65) {$\Theta_c$};
		\node [style=none] (31) at (5.225, 4.85) {};
		\node [style=none] (32) at (5.225, 1.7) {};
	\end{pgfonlayer}
	\begin{pgfonlayer}{edgelayer}
		\draw [in=165, out=0, looseness=1.25] (2) to (0);
		\draw [in=-180, out=0, looseness=1.25] (1) to (0);
		\draw [in=90, out=0] (4.center) to (6.center);
		\draw [in=-90, out=0] (5.center) to (6.center);
		\draw (0) to (4.center);
		\draw [in=90, out=180] (7.center) to (9.center);
		\draw [in=-90, out=-180] (8.center) to (9.center);
		\draw (8.center) to (2);
		\draw [style=norm] (10.center) to (11.center);
		\draw (7.center) to (14.center);
		\draw [style=norm] (31.center) to (32.center);
		\draw [style=norm] (11.center) to (32.center);
		\draw [style=norm] (10.center) to (31.center);
		\draw (3) to (5.center);
	\end{pgfonlayer}
\end{tikzpicture}

}

\newcommand{\trainingSequentialY}{
\begin{tikzpicture}
	\begin{pgfonlayer}{nodelayer}
		\node [style=small box] (0) at (5.05, 3.25) {$\pdist_{\rv{C}_\rv{T} \mid \rv{X}; \Theta_c}$};
		\node [style=triangle] (1) at (3.525, 2.75) {$x$};
		\node [style=triangle] (2) at (3.525, 3.725) {$\theta_c$};
		\node [style=small box] (19) at (8.425, 3.25) {$\pdist_{\rv{Y}_\rv{T} \mid \rv{C}; \Theta_y}$};
		\node [style=small box] (21) at (7, 3.7) {$\pdist_{\Theta_y}$};
		\node [style=triangle] (22) at (3.525, 1.85) {$y$};
		\node [style=none] (23) at (9.25, 3.25) {};
		\node [style=none] (24) at (9.25, 1.85) {};
		\node [style=none] (25) at (9.65, 2.55) {};
		\node [style=none] (26) at (6.475, 4.2) {};
		\node [style=none] (27) at (6.475, 3.7) {};
		\node [style=none] (28) at (6.125, 3.95) {};
		\node [style=none] (33) at (10.4, 4.2) {};
		\node [style=none] (34) at (10.725, 4.2) {$\Theta_y$};
		\node [style=none] (39) at (3.975, 4.35) {};
		\node [style=none] (40) at (3.975, 1.675) {};
		\node [style=none] (41) at (9.975, 4.35) {};
		\node [style=none] (42) at (9.975, 1.675) {};
	\end{pgfonlayer}
	\begin{pgfonlayer}{edgelayer}
		\draw [in=165, out=0, looseness=1.25] (2) to (0);
		\draw [in=-165, out=0, looseness=1.25] (1) to (0);
		\draw [in=165, out=0, looseness=1.25] (21) to (19);
		\draw [in=90, out=0] (23.center) to (25.center);
		\draw [in=-90, out=0] (24.center) to (25.center);
		\draw (19) to (23.center);
		\draw [in=90, out=180] (26.center) to (28.center);
		\draw [in=-90, out=-180] (27.center) to (28.center);
		\draw (27.center) to (21);
		\draw (26.center) to (33.center);
		\draw [style=norm] (39.center) to (40.center);
		\draw [style=norm] (41.center) to (42.center);
		\draw [style=norm] (40.center) to (42.center);
		\draw [style=norm] (39.center) to (41.center);
		\draw (22) to (24.center);
		\draw (0) to (19);
	\end{pgfonlayer}
\end{tikzpicture}

}

\newcommand{\trainingSequentialYMAPC}{
\begin{tikzpicture}
	\begin{pgfonlayer}{nodelayer}
		\node [style=small box] (0) at (5.05, 3.25) {$\pdist_{\rv{C}_\rv{T} \mid \rv{X}; \Theta_c}$};
		\node [style=triangle] (1) at (3.525, 2.75) {$x$};
		\node [style=triangle] (2) at (3.525, 3.725) {$\theta_c$};
		\node [style=small box] (19) at (8.675, 3.25) {$\pdist_{\rv{Y}_\rv{T} \mid \rv{C}; \Theta_y}$};
		\node [style=small box] (21) at (7.25, 3.7) {$\pdist_{\Theta_y}$};
		\node [style=triangle] (22) at (3.525, 1.85) {$y$};
		\node [style=none] (23) at (9.5, 3.25) {};
		\node [style=none] (24) at (9.5, 1.85) {};
		\node [style=none] (25) at (9.9, 2.55) {};
		\node [style=none] (26) at (6.725, 4.2) {};
		\node [style=none] (27) at (6.725, 3.7) {};
		\node [style=none] (28) at (6.375, 3.95) {};
		\node [style=none] (33) at (10.65, 4.2) {};
		\node [style=none] (34) at (10.975, 4.2) {$\Theta_y$};
		\node [style=none] (39) at (6.225, 4.35) {};
		\node [style=none] (40) at (6.225, 1.675) {};
		\node [style=none] (41) at (10.225, 4.35) {};
		\node [style=none] (42) at (10.225, 1.675) {};
		\node [style=none] (43) at (3.975, 3.85) {};
		\node [style=none] (44) at (6.05, 3.85) {};
		\node [style=none] (45) at (3.975, 2.6) {};
		\node [style=none] (46) at (6.05, 2.6) {};
	\end{pgfonlayer}
	\begin{pgfonlayer}{edgelayer}
		\draw [in=165, out=0, looseness=1.25] (2) to (0);
		\draw [in=-165, out=0, looseness=1.25] (1) to (0);
		\draw [in=165, out=0, looseness=1.25] (21) to (19);
		\draw [in=90, out=0] (23.center) to (25.center);
		\draw [in=-90, out=0] (24.center) to (25.center);
		\draw (19) to (23.center);
		\draw [in=90, out=180] (26.center) to (28.center);
		\draw [in=-90, out=-180] (27.center) to (28.center);
		\draw (27.center) to (21);
		\draw (26.center) to (33.center);
		\draw [style=norm] (39.center) to (40.center);
		\draw [style=norm] (41.center) to (42.center);
		\draw [style=norm] (40.center) to (42.center);
		\draw [style=norm] (39.center) to (41.center);
		\draw (22) to (24.center);
		\draw (0) to (19);
		\draw [style=amax] (43.center) to (44.center);
		\draw [style=amax] (44.center) to (46.center);
		\draw [style=amax] (46.center) to (45.center);
		\draw [style=amax] (45.center) to (43.center);
	\end{pgfonlayer}
\end{tikzpicture}

}

\newcommand{\trainingJoint}{
\begin{tikzpicture}
	\begin{pgfonlayer}{nodelayer}
		\node [style=small box] (0) at (5.425, 3.25) {$\pdist_{\rv{C}_\rv{T} \mid \rv{X}; \Theta_c}$};
		\node [style=triangle] (1) at (2.475, 2.925) {$x$};
		\node [style=small box] (2) at (4, 3.7) {$\pdist_{\Theta_c}$};
		\node [style=none] (7) at (3.475, 4.65) {};
		\node [style=none] (8) at (3.475, 3.7) {};
		\node [style=none] (9) at (3.125, 4.175) {};
		\node [style=none] (10) at (2.975, 4.85) {};
		\node [style=none] (11) at (2.975, 1.7) {};
		\node [style=none] (14) at (10.4, 4.65) {};
		\node [style=none] (15) at (10.725, 4.65) {$\Theta_c$};
		\node [style=small box] (19) at (8.425, 3.25) {$\pdist_{\rv{Y}_\rv{T} \mid \rv{C}; \Theta_y}$};
		\node [style=small box] (21) at (7, 3.7) {$\pdist_{\Theta_y}$};
		\node [style=triangle] (22) at (2.475, 1.95) {$y$};
		\node [style=none] (23) at (9.25, 3.25) {};
		\node [style=none] (24) at (9.25, 1.95) {};
		\node [style=none] (25) at (9.65, 2.6) {};
		\node [style=none] (26) at (6.475, 4.2) {};
		\node [style=none] (27) at (6.475, 3.7) {};
		\node [style=none] (28) at (6.125, 3.95) {};
		\node [style=none] (31) at (9.975, 4.85) {};
		\node [style=none] (32) at (9.975, 1.7) {};
		\node [style=none] (33) at (10.4, 4.2) {};
		\node [style=none] (34) at (10.725, 4.2) {$\Theta_y$};
	\end{pgfonlayer}
	\begin{pgfonlayer}{edgelayer}
		\draw [in=165, out=0, looseness=1.25] (2) to (0);
		\draw [in=-180, out=0, looseness=1.25] (1) to (0);
		\draw [in=90, out=180] (7.center) to (9.center);
		\draw [in=-90, out=-180] (8.center) to (9.center);
		\draw (8.center) to (2);
		\draw [style=norm] (10.center) to (11.center);
		\draw (7.center) to (14.center);
		\draw [in=165, out=0, looseness=1.25] (21) to (19);
		\draw [in=90, out=0] (23.center) to (25.center);
		\draw [in=-90, out=0] (24.center) to (25.center);
		\draw (19) to (23.center);
		\draw [in=90, out=180] (26.center) to (28.center);
		\draw [in=-90, out=-180] (27.center) to (28.center);
		\draw (27.center) to (21);
		\draw [style=norm] (31.center) to (32.center);
		\draw (26.center) to (33.center);
		\draw [style=norm] (11.center) to (32.center);
		\draw [style=norm] (10.center) to (31.center);
		\draw (0) to (19);
		\draw (22) to (24.center);
	\end{pgfonlayer}
\end{tikzpicture}

}

\newcommand{\trainingJointMAPC}{
\begin{tikzpicture}
	\begin{pgfonlayer}{nodelayer}
		\node [style=small box] (0) at (5.425, 3.25) {$\pdist_{\rv{C}_\rv{T} \mid \rv{X}; \Theta_c}$};
		\node [style=triangle] (1) at (2.225, 2.925) {$x$};
		\node [style=small box] (2) at (4, 3.7) {$\pdist_{\Theta_c}$};
		\node [style=none] (7) at (3.475, 4.65) {};
		\node [style=none] (8) at (3.475, 3.7) {};
		\node [style=none] (9) at (3.125, 4.175) {};
		\node [style=none] (10) at (2.725, 5.1) {};
		\node [style=none] (11) at (2.725, 1.7) {};
		\node [style=none] (14) at (10.9, 4.65) {};
		\node [style=none] (15) at (11.225, 4.65) {$\Theta_c$};
		\node [style=small box] (19) at (8.925, 3.25) {$\pdist_{\rv{Y}_\rv{T} \mid \rv{C}; \Theta_y}$};
		\node [style=small box] (21) at (7.5, 3.7) {$\pdist_{\Theta_y}$};
		\node [style=triangle] (22) at (2.225, 1.95) {$y$};
		\node [style=none] (23) at (9.75, 3.25) {};
		\node [style=none] (24) at (9.75, 1.95) {};
		\node [style=none] (25) at (10.15, 2.6) {};
		\node [style=none] (26) at (6.975, 4.2) {};
		\node [style=none] (27) at (6.975, 3.7) {};
		\node [style=none] (28) at (6.625, 3.95) {};
		\node [style=none] (31) at (10.475, 5.1) {};
		\node [style=none] (32) at (10.475, 1.7) {};
		\node [style=none] (33) at (10.9, 4.2) {};
		\node [style=none] (34) at (11.225, 4.2) {$\Theta_y$};
		\node [style=none] (35) at (2.875, 4.85) {};
		\node [style=none] (36) at (6.475, 4.85) {};
		\node [style=none] (37) at (6.475, 2.85) {};
		\node [style=none] (38) at (2.875, 2.85) {};
	\end{pgfonlayer}
	\begin{pgfonlayer}{edgelayer}
		\draw [in=165, out=0, looseness=1.25] (2) to (0);
		\draw [in=-180, out=0, looseness=1.25] (1) to (0);
		\draw [in=90, out=180] (7.center) to (9.center);
		\draw [in=-90, out=-180] (8.center) to (9.center);
		\draw (8.center) to (2);
		\draw [style=norm] (10.center) to (11.center);
		\draw (7.center) to (14.center);
		\draw [in=165, out=0, looseness=1.25] (21) to (19);
		\draw [in=90, out=0] (23.center) to (25.center);
		\draw [in=-90, out=0] (24.center) to (25.center);
		\draw (19) to (23.center);
		\draw [in=90, out=180] (26.center) to (28.center);
		\draw [in=-90, out=-180] (27.center) to (28.center);
		\draw (27.center) to (21);
		\draw [style=norm] (31.center) to (32.center);
		\draw (26.center) to (33.center);
		\draw [style=norm] (11.center) to (32.center);
		\draw [style=norm] (10.center) to (31.center);
		\draw (0) to (19);
		\draw (22) to (24.center);
		\draw [style=amax] (35.center) to (38.center);
		\draw [style=amax] (38.center) to (37.center);
		\draw [style=amax] (37.center) to (36.center);
		\draw [style=amax] (35.center) to (36.center);
	\end{pgfonlayer}
\end{tikzpicture}

}

\newcommand{\conceptProcess}{
\begin{tikzpicture}
	\begin{pgfonlayer}{nodelayer}
		\node [style=small box] (1) at (4.175, 3.25) {$\pdist_{\rv{C} \mid \text{pa}(\rv{C})}$};
		\node [style=none] (3) at (5.25, 3.25) {};
		\node [style=none] (4) at (5.5, 3.25) {$\rv{C}$};
		\node [style=none] (5) at (3, 3.25) {};
		\node [style=none] (6) at (2.5, 3.25) {$\text{pa}(\rv{C})$};
	\end{pgfonlayer}
	\begin{pgfonlayer}{edgelayer}
		\draw (1) to (3.center);
		\draw (5.center) to (1);
	\end{pgfonlayer}
\end{tikzpicture}

}

\newcommand{\inlineimg}[1]{%
  \smash{\raisebox{-.2\height}{\includegraphics[height=.9em]{figs/#1.png}}}%
}

\icmltitlerunning{Actionable Interpretability Must Be Defined in Terms of Symmetries}

\usepackage[dvipsnames]{xcolor}

\definecolor{softblue}{RGB}{173, 216, 230} 
\definecolor{softgreen}{RGB}{144, 238, 144} 
\definecolor{softpink}{RGB}{255, 182, 193} 
\definecolor{softpurple}{RGB}{221, 160, 221} 
\definecolor{softorange}{RGB}{255, 224, 178} 
\definecolor{mediumblue}{RGB}{100, 149, 237} 
\definecolor{mediumred}{RGB}{220, 60, 60} 

\usepackage{enumitem}

\newcommand{\mez}[1]{\textcolor{orange}{[MEZ] #1}}

\newcommand{\todo}[1]{\textcolor{red}{[TODO] #1}}

\usepackage{adjustbox}
\usepackage{comment}

\usepackage{tikz}
\usepackage{tikz-cd}
\usetikzlibrary{positioning, arrows.meta, shapes}
\usetikzlibrary{arrows.meta, positioning, quotes, fit, shapes, backgrounds, calc}

\DeclareMathOperator*{\argmax}{arg\,max}

\newcommand{\rv}[1]{\MakeUppercase{#1}}
\newcommand{\myset}[1]{\myset{#1}}
\newcommand{\s}[1]{\mathcal{\MakeUppercase{#1}}}

\def\cidx{i}

\def\P{\mathbb{P}}
\newcommand{\Prob}[1]{\P\left( {#1} \right)}

\def\pdist{P}
\newcommand{\cat}[1]{\mathsf{#1}}

\begin{document}

\twocolumn[
\icmltitle{Actionable Interpretability Must Be Defined in Terms of Symmetries}



\icmlsetsymbol{equal}{*}


\begin{icmlauthorlist}
\icmlauthor{Pietro Barbiero}{ibm}
\icmlauthor{Mateo Espinosa Zarlenga}{ox}
\icmlauthor{Francesco Giannini}{unipi}
\icmlauthor{Alberto Termine}{supsi}
\icmlauthor{Filippo Bonchi}{unipi}
\icmlauthor{Mateja Jamnik}{cam}
\icmlauthor{Giuseppe Marra}{ku}
\end{icmlauthorlist}

\icmlaffiliation{ibm}{IBM Research}
\icmlaffiliation{ox}{University of Oxford}
\icmlaffiliation{unipi}{University of Pisa}
\icmlaffiliation{supsi}{SUPSI}
\icmlaffiliation{cam}{University of Cambridge}
\icmlaffiliation{ku}{KU Leuven}

\icmlcorrespondingauthor{Pietro Barbiero}{barbiero@tutanota.com}

\icmlkeywords{Machine Learning, ICML}

\vskip 0.3in
]



\printAffiliationsAndNotice{} 

\begin{abstract}
This paper argues that interpretability research in Artificial Intelligence (AI) is fundamentally ill-posed as existing definitions of interpretability fail to describe how interpretability can be formally tested or designed for. 
We posit that actionable definitions of interpretability must be formulated in terms of \emph{symmetries} that inform model design and lead to testable conditions.
Under a \emph{compositional} view, we hypothesise that four symmetries (inference equivariance, information invariance, concept-closure invariance, and structural invariance) suffice to (i)~formalise interpretable models as a subclass of probabilistic models, (ii) yield a unified formulation of interpretable inference (e.g., alignment, interventions, and counterfactuals) as a form of Bayesian inversion, and (iii) provide a formal framework to verify compliance with safety standards and regulations.\looseness-1
\end{abstract}

\section{Introduction}
Recent years have seen a surge in interpretable models whose decisions are easy for humans to understand. These models now offer a performance comparable to that of powerful black-box models like Deep Neural Networks (DNNs)~\citep{senn,protopnet,cem}, and are increasingly employed to diagnose errors, ensure fairness, and attain legal compliance~\citep{lee2021development, meng2022interpretability}.

While significant, we argue that current research in interpretable Artificial Intelligence (AI) is \textit{ill-posed}. We claim this is the case since existing foundational efforts (Sec.~\ref{sec:alternative}), such as those by \citet{kim2016examples}, \citet{biran2017explanation} \citet{doshi2017towards}, \citet{lipton2018mythos}, \citet{miller2019explanation}, \citet{watson2021explanation}, \citet{facchini2021towards}, \citet{giannini2024categorical}, and \citet{tull2024towards}, all fail to provide (1) a formal framework for verifying interpretability and (2) a set of design principles for designing interpretable models.


In the spirit of the Erlangen Program \citep{klein1893comparative,bronstein2021geometric}, in this paper \textbf{we posit that an actionable definition of interpretability must be given in terms of \emph{symmetries}, that is, structure-preserving transformations.} 
We argue that four symmetries suffice to capture the essence of interpretability across the literature (Sec.~\ref{sec:invariances}):
\begin{itemize}[topsep=0pt, leftmargin=9pt, itemsep=-2pt]
    \item \textbf{Inference equivariance.}
    A model is interpretable if a user can correctly predict the model outputs.

    \item \textbf{Information invariance.}  
    An interpretable model should retain only the input information that is sufficient for the task, discarding irrelevant details (e.g., an individual pixel intensity is unnecessary for classifying cats vs.\ dogs).

    \item \textbf{Concept-closure invariance.}  
    The variables used by an interpretable model should have the same semantics as those used by humans (e.g., a concept ``\textit{red}'' in the model matches ``\textit{red}'' for a human).

    \item \textbf{Structural invariance.}  
    A model is interpretable if it is drawn from a hypothesis class that the target user can reason about. For example, if a user can only reason in linear terms, the model must also be linear.

\end{itemize}

A definition of interpretability satisfying these symmetries is \emph{actionable} since they (i)~make interpretability testable by enabling the formal verification of whether, and to what extent, a model satisfies each symmetry, (ii)~inform model design by identifying which symmetries are violated, and (iii)~subsume interpretability properties studied in the literature as special cases.
Under a compositional probabilistic view, we show how these symmetries allow us to formalise interpretable models as a Markov category, that is, a recipe specifying the ingredients and the elementary cooking operations to build any interpretable model (Sec.~\ref{sec:models}). Finally, we show how the constructed category unifies alignment, intervention, and counterfactual inference as a form of Bayesian inversion (Sec.~\ref{sec:inference}). For an overview of our position, the main ideas of the paper can be understood in under three minutes by reading only the setup and the highlighted text boxes.

\section{Interpretability Symmetries}
\label{sec:invariances}

In this section, we introduce four symmetries and hypothesise that their satisfaction characterises the class of models that are interpretable to a given human user. Inspired by mathematical foundations of explainable AI \cite{giannini2024categorical,tull2024towards,colombini2025mathematical,fioravanti2025categorical}, we rely on category theory\footnote{While not essential to understand this paper, basic notions in category theory related to this work are presented in \cite{jacobs2025compositional,tull2024towards}.} as an expressive framework to characterise (i)~human users and (ii)~the class of models that are interpretable.
\paragraph{Setup.}

Following \citet{murphy2023probabilistic}, we adopt a probabilistic perspective. From a categorical standpoint, the relevant structures are Markov categories \citep{fritz2020synthetic}, which provide a general and compositional language for probabilistic modelling. In what follows, we focus on the category 
$\cat{BorelStoch}$ of Standard Borel spaces and Markov kernels \cite{lawvere1962category,giry2006categorical,panangaden2009labelled,fritz2020synthetic}; however, analogous arguments can be developed in an arbitrary Markov category.
\textbf{Human:} We denote by $h$ a human user and by $\cat{Hm}$ the category of models (i.e., the hypothesis space) of the user's ``\emph{mental models}'' \citep{johnson1983mental,kulesza2013too,kim2018interpretability}. 
Formally, $\cat{Hm}$ is characterised by a set of user-dependent structural properties (e.g., linearity, sparsity, or monotonicity) which are assumed to be given.
\textbf{Model:} We then define $\cat{Im}_{[h]}$ as the category of models that are interpretable for $h$, and consider both $\cat{Hm}$ and $\cat{Im}_{[h]}$ to be sub-Markov categories of $\cat{Stoch}$ (formally, there exists faithful embeddings of Markov categories $E_1: \cat{Im} \hookrightarrow \cat{BorelStoch}$ and $E_2: \cat{Hm} \hookrightarrow \cat{BorelStoch}$).
\textbf{Notation:} Let $\Omega$ be an underlying sample space. A random variable $\rv{X}:\Omega\to\s{X}$ maps each outcome $\omega\in\Omega$ to a feature representation $x=\rv{X}(\omega)\in\s{X}$ (e.g., pixels), and a random variable $\rv{Y}:\Omega\to\s{Y}$ maps outcomes to task-relevant labels (e.g., animal species). A probabilistic model is a conditional distribution $\pdist_{\rv{Y}\mid\rv{X}}:\s{X}\to\Prob{\s{Y}}$ that assigns to each feature value $x\in\s{X}$ a probability measure $\pdist_{\rv{Y}\mid\rv{X}}(x)=P(\rv{Y}\in\cdot\mid\rv{X}=x)$ on $\s{Y}$; in categorical probability, this is a stochastic morphism $\s{X}\xrightarrow[]{\pdist_{\rv{Y} \mid \rv{X}}}\s{Y}$. Probabilistic inference consists of evaluating $\pdist_{\rv{Y}\mid\rv{X}}(x)$ for a given input $x$ to obtain a predictive distribution over $\s{Y}$.
To rule out trivial models, we are interested in models where $\rv{Y}$ depends on more than one component of $\rv{X}=(\rv{X}_1,\dots,\rv{X}_n)$, that is, $\pdist_{\rv{Y}\mid\rv{X}} \neq \pdist_{\rv{X}_i\mid\rv{X}}$.



\subsection{Symmetry I: Inference Equivariance}
Most works in the interpretability literature define their subject matter informally. For instance, \citet{kim2016examples} and \citet{miller2019explanation} suggest that \emph{a method is interpretable if a user can correctly and efficiently predict the method's results} and \emph{the cause of such results}. Closely related, \citet{biran2017explanation} and \citet{murdoch2019interpretable} argue that a \emph{system is interpretable if a human is able to internally simulate and reason about its entire decision-making process (i.e., how a trained model produces an output for an arbitrary input)}. Similar views connecting the notion of interpretability to how \textit{intelligible} a model's inference is have been proposed by \citet{lou2012intelligible} and \citet{lime}. While influential, these definitions remain largely descriptive. Our first goal is then to answer the following question: \noindent \textbf{(RQ1)}~\emph{How can we formally bring together the informal descriptions of interpretability?}\looseness-1

To answer this question, we introduce a concrete example showing what it would mean for a user to ``simulate a model'' and ``predict the model's results''. Consider a finite space of values for $\rv{X}$ and $\rv{Y}$, with $\rv{X}=(\rv{X}_1,\rv{X}_2)$, so that we can give a complete representation of the model's behaviour via a conditional probability table:
\begin{center}
    \vspace*{-2mm}
    \text{\textbf{Table 1}: Tabular representation of a model $\pdist_{\rv{Y} \mid \rv{X}}$.}
    \label{eq:cpt}
    \vspace{-1.5em}
\end{center}
\begin{align*} 
{\begin{array}{c|cc|c}
 & \rv{X}_1 & \rv{X}_2
 & P(\rv{Y}=1 \mid \rv{X}_1=x_1,\; \rv{X}_2=x_2) \\
\hline
\inlineimg{zeroblack} & 0 & 1 & 1 \\
\inlineimg{zerowhite} & 0 & 0 & 1 \\
\inlineimg{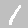} & 1 & 0 & 0 \\
\end{array}}
\end{align*}
We now fix a \emph{translation map} that associates the model's objects with representations meaningful to a given user $h$, that is, $\tau_\s{X} : \s{X} \to \s{X}_{[h]}$ and $\tau_\s{Y} : \s{Y} \to \s{Y}_{[h]}$. For example, $\tau_\s{X}$ may map the objects $\s{X}_1 \times \s{X}_2$ to $\s{X}_{\texttt{one}} \times \s{X}_{\texttt{black}}$. Given the original model and the translation maps $\tau$, the user $h$ can ``predict the model's results'' if the user can build a ``mental model'' $\pdist^{[h]}_{\rv{Y} \mid \rv{X}}$ in $\cat{Hm}$ whose predictions match the translated outputs of $\pdist_{\rv{Y} \mid \rv{X}}$.
In practice, this is possible if the following procedures are equivalent: (1)~we can first translate the input features $\rv{X}$ into a human input space $\rv{X}_{[h]}$ and predict the target using the human model $\pdist^{[h]}_{\rv{Y} \mid \rv{X}}$, or (2)~we can first infer a label using $\pdist_{\rv{Y} \mid \rv{X}}$ and then translate the result. 
Visually, we represent these two paths as:
\[{\small 
\begin{adjustbox}{max width=\linewidth}
\begin{tikzcd}[column sep=3.5cm, row sep=normal]
(X_1, X_2) = (0, 0) \arrow[r, "\text{unknown model } \pdist_{\rv{Y} \mid \rv{X}}"] \arrow[d, "\text{``translate''}"'] & Y=1 \arrow[d, "\text{``translate''}"] \\
{\scriptscriptstyle
\shortstack{
$(\rv{X}_\texttt{one}, \rv{X}_\texttt{black}) =$
$(\text{no}, \text{no})$
}
} \arrow[r, "\text{human model } \pdist^{[h]}_{\rv{Y} \mid \rv{X}}"']  & \rv{Y}_\texttt{even}=\text{yes}
\end{tikzcd}
\end{adjustbox}
}\]
If this diagram commutes for \textit{any} input $x \in \s{X}$ (i.e., if we reach the same result following different paths), then it means that the outputs of the model $\pdist_{\rv{Y} \mid \rv{X}}$ can be obtained using the human model $\pdist^{[h]}_{\rv{Y} \mid \rv{X}}$. Inspired by \citet{marconato2023interpretability}, we refer to this principle as \textit{inference equivariance}:

\smallskip
\begin{definition3}(\textbf{Inference Equivariance}) 
\label{def:sym-one}
    Inference $\pdist_{\rv{Y} \mid \rv{X}}(\rv{Y} \mid \rv{X} = x)$ is \emph{equivariant} w.r.t.\ a reference $\cat{Hm}$ under translations $\tau_\s{X} : \s{X} \to \s{X}_{[h]}$ and $\tau_\s{Y} : \s{Y} \to \s{Y}_{[h]}$ if and only if there exists a model $\pdist^{[h]}_{\rv{Y} \mid \rv{X}}: \s{X}_{[h]} \to \Prob{\s{Y}}$ in $\cat{Hm}$ such that the following diagram commutes:
    \vspace*{-2mm}
    \[%
        {
            \begin{adjustbox}{max width=\linewidth}
            \begin{tikzcd}[column sep=1.5cm, row sep=normal]
            \s{X} \arrow[r, "\pdist_{\rv{Y} \mid \rv{X}}"] \arrow[d, "\tau_\s{X}"'] & \s{Y} \arrow[d, "\tau_\s{Y}"] \\
            \s{X}_{[h]} \arrow[r, "\pdist^{[h]}_{\rv{Y} \mid \rv{X}}"']  & \s{Y}_{[h]}
            \end{tikzcd}
            \end{adjustbox}
        }%
    \]
\end{definition3}



 \subsubsection{Consequences of Inference Equivariance}
 While this symmetry subsumes prior definitions of interpretability by \citet{kim2016examples}, \citet{miller2019explanation}, \citet{biran2017explanation}, and \citet{murdoch2019interpretable}, it leaves several important \textit{challenges} open: 
\begin{enumerate}[topsep=0pt, leftmargin=18pt, itemsep=-2pt]
    \item[\textbf{C1}] 
    \textbf{Naively verifying inference equivariance is intractable.}
    To guarantee that inference equivariance always holds for any $x \in \s{X}$, we need a table with $\mathcal{O}(\exp(|\s{X}|))$ entries. Hence, if we consider even small binary images whose pixel space is $\{0,1\}^{10 \times 10}$, we would already need more evaluations than the number of atoms in the observable universe.
    \item[\textbf{C2}] 
    \textbf{Many translations exist, but some are not sound.} As we will discuss later, not all translations lead to commutative diagrams. Therefore, it is important to consider how to construct such a translation. 
    \item[\textbf{C3}] 
    \textbf{Many models $\pdist_{\rv{Y} \mid \rv{X}}$ may exist, but some may not satisfy desirable user-specific properties.} Hence, simply verifying inference equivariance does not mean that users understand the internal behaviour of $\pdist_{\rv{Y} \mid \rv{X}}$.
\end{enumerate}
Therefore, we can conclude the following:
\smallskip
\begin{definition4}
A user's ability to predict a model's output is essential (but not enough) for the model $\pdist_{\rv{Y} \mid \rv{X}}$ to be interpretable.
\end{definition4}
Together, the challenges outlined above highlight the limits of informal definitions in the literature; that is, without a formal framework, key limitations are more easily overlooked. In the following sections, we address each challenge by introducing a corresponding symmetry. 


\subsection{Symmetry II: Information Invariance}
\label{sec:factorisation}
We first tackle challenge \textbf{C1}, that is the fact that verifying inference equivariance is intractable in large input spaces. With this challenge in mind, our first goal is then to answer:
\textbf{(RQ2)}~\emph{How can we make inference equivariance tractable?}

Ideally, one would like to work with a model $\pdist_{\rv{Y} \mid \rv{X}}$ for which inference equivariance can be verified in only a few steps. When this is not possible, we need to rewrite $\pdist_{\rv{Y} \mid \rv{X}}$ to make inference equivariance more tractable.
This requires a form of compression which captures \emph{only the essential properties} of each input $x \in \s{X}$ related to $\rv{Y}$.
To this end, we introduce a surjective map $\pdist_{\rv{Z} \mid \rv{X}}: \s{X} \twoheadrightarrow \Prob{\s{Z}}$ that, when composed with a model $\pdist_{\rv{Y} \mid \rv{Z}}$, enables us to obtain $\pdist_{\rv{Y} \mid \rv{X}}$ again. 
The effect of this transformation can be characterised by analysing the information content in $\rv{Z}$ compared to $\rv{X}$ (quantified by entropy $H(\cdot)$) and by analysing how much information we have left in $\rv{Z}$ to predict $\rv{Y}$ compared to $\rv{X}$ (quantified by the mutual information $I(\cdot ; \cdot)$). Specifically, we argue that \textbf{C1} can be circumvented if the surjection $\pdist_{\rv{Z} \mid \rv{X}}$ satisfies the following invariance:\looseness-1

\smallskip
\begin{definition3}\textbf{(Information Invariance)}
\label{def:sym-two}
    Given a model $\pdist_{\rv{Y} \mid \rv{X}}$, the mutual information $I(\rv{Y};\rv{X})$ is invariant under marginalisation $\pdist_{\rv{Y} \mid \rv{Z}} \circ \pdist_{\rv{Z} \mid \rv{X}}$ with $H(\rv{Z}) \ll H(\rv{X})$ if the following diagram commutes:\vspace*{-3mm}
        \[
        \begin{tikzcd}[column sep=normal, row sep=normal]
        \s{X} \arrow[r, two heads, "\pdist_{\rv{Z} \mid \rv{X}}"] \arrow[rd, "\pdist_{\rv{Y} \mid \rv{X}}"'] & \s{Z} \arrow[d, "\pdist_{\rv{Y} \mid \rv{Z}}"] \\
        & \s{Y}
        \end{tikzcd}
        \]\vspace*{-5mm}
\end{definition3}

This invariance guarantees that $\rv{Z}$ retains all information in $\rv{X}$ that is relevant for predicting $\rv{Y}$ while discarding extraneous variability. 
\begin{remark}
    The marginalisation $\pdist_{\rv{Y} \mid \rv{Z}} \circ \pdist_{\rv{Z} \mid \rv{X}}$ makes the variable $\rv{Y}$ \textbf{conditionally independent} with respect to $\rv{X}$ given $\rv{Z}$, that is, $I(\rv{Y}; \rv{X} \mid \rv{Z}) = 0$.
\end{remark}
If $\rv{Y}$ is conditionally independent with respect to $\rv{X}$, then $\rv{X}$ does not provide any information to explain $\rv{Y}$ once we know $\rv{Z}$. So, when verifying inference equivariance for a model $\pdist_{\rv{Y} \mid \rv{Z}}$, we can safely ignore $\rv{X}$. Since $H(\rv{Z}) \ll H(\rv{X})$ by construction, the conditional probability table representing $\pdist_{\rv{Y} \mid \rv{Z}}$ is exponentially smaller than that for $\pdist_{\rv{Y} \mid \rv{X}}$. This leads to the following conclusion:

\smallskip
\begin{definition4}
Verifying inference equivariance for $\pdist_{\rv{Y} \mid \rv{Z}}$ is equivalent but more tractable than for $\pdist_{\rv{Y} \mid \rv{X}}$.
\end{definition4}

\subsubsection{Consequences of Information Invariance}
Requiring interpretable models to satisfy information invariance allows us to derive the following consequences:

\myparagraph{Derivable properties.} 
Properties that are lauded as desirable in the interpretability literature, such as \textit{sparsity} (few features, few parameters), \textit{compactness} (exclusion of irrelevant information) \citep{murphy2023probabilistic}, \textit{completeness} (explanations are sufficient statistics for model prediction) \citep{yeh2020completeness}, and \textit{modularity} (a model can be broken down into simpler components) \citep{murdoch2019interpretable},  can be seen as mechanisms that enable information invariance rather than being primitive notions.

\myparagraph{Methods to guarantee information invariance.}
Models employing feature selection \citep{miller1984selection} (e.g., sparse decision trees) or leveraging the manifold hypothesis \citep{bengio2013representation} (e.g., DNNs \citep{tishby2015deep}) can enforce information invariance by design.

\myparagraph{Disqualified approaches.}
Traditional feature-attribution \citep{lime, shap, og_saliency, integrated_gradients} operates in the original input space and does not guarantee the existence of a lower-dimensional representation $\rv{Z}$ that captures all and only the information relevant for $\rv{Y}$. Therefore, by failing to satisfy information invariance, verifying inference equivariance using post-hoc feature attribution remains intractable.


\subsection{Symmetry III: Concept Closure Invariance}
\label{sec:concept}

\Cref{def:sym-two} lets us verify inference equivariance based on $\pdist_{\rv{Y} \mid \rv{Z}}$. However, as framed by challenge \textbf{C2}, it does not give us a criterion for characterising translations that make the inference diagram commute.
Therefore, we study what makes a translation \textit{sound}, in the sense that it that preserves ``meaning''.
This leads to the following research question:
\textbf{(RQ3)}~\emph{What is required for a translation to be sound?}


To address this question, we first introduce a key data structure, namely \emph{concepts}, which allows us to characterise sound translations. Following \citet{goguen2005concept} and \citet{ganter1996formal}, we think of a concept as a relation between a set of 
objects (e.g., $\{\inlineimg{oneblack}, \inlineimg{zeroblack}, \inlineimg{appleb}, \dots\}$) and  symbols (e.g., $\{\texttt{black},\texttt{round},\ldots\}$) as shown in the following example:

\begin{example}
    Consider a set of sentences $\s{S}=\{ \texttt{black},$ $ \texttt{white},$ $\texttt{round},$ $\texttt{even} \}$ and a set of objects $\s{U}$.
     Here, we can fully define what we mean by the concept ``black'' via the tuple $(\s{T},\s{M})$ with  $\s{T}=\{\texttt{black}\} \subseteq \s{S}$ and  
$\s{M}=\{ \inlineimg{appleb}, \inlineimg{oneblack}, \inlineimg{zeroblack} \}\subseteq\s{U}$, as shown in the next figure: 
\begin{align*}
        \resizebox{.68\columnwidth}{!}{\exampleConceptRound}
\vspace*{-2mm}
\end{align*}
\end{example}
Given a set of objects $\s{U}$ and a set of sentences $\s{S}$, we can formalise the notion of a concept captured by the example above by considering the functions $\beta : \mathcal{P}(\s{S}) \to \mathcal{P}(\s{U})$, which maps any $\s{T} \subseteq \s{S}$ into the subset of objects of $\s{U}$ whose elements satisfy every sentence $\varphi$ in $\s{T}$, and $\gamma : \mathcal{P}(\s{U}) \to \mathcal{P}(\s{S})$, which maps any $\s{M} \subseteq \s{U}$ into the subset of all sentences in $\s{S}$ satisfied by every object in $\s{M}$.
In this context, a concept can be defined as follows:

\smallskip
\begin{definition2}[Concept]
Given objects $\s{U}$, sentences $\s{S}$, and functions $\beta,\gamma$ as above, a \emph{concept} is a tuple $(\s{T},\s{M}) \in \mathcal{P}(\s{S})\times\mathcal{P}(\s{U})$ representing a fixed point of $\beta$ and $\gamma$, that is, 
$\s{T} = \gamma(\s{M}) \text{ and } \s{M} = \beta(\s{T})$.
\end{definition2}    

Under this definition, we characterise a \textit{sound translation} as a \textit{concept-preserving} map associating different symbols (e.g., \texttt{black} and \texttt{noir}) to the same objects (e.g., $\{\inlineimg{oneblack}, \inlineimg{zeroblack}, \inlineimg{appleb}, \dots\}$). Intuitively, concepts allow the characterisation of sound translations, that is, translations that ``preserve concepts'', by noticing that, if an object satisfies a sentence $\varphi$, it should also satisfy the translated sentence $\tau(\varphi)$. We refer to such sound translations as \textit{concept-based translations} $\tau_\s{C}$.
Concept-based translations are sound, and therefore enable interpretability via inference equivariance, if they satisfy the following invariance:

\smallskip
\begin{definition3}\textbf{(Concept Closure Invariance)}
\label{def:sym-three}
Concept closure is invariant under a sentence translation function $\tau_\s{C} : \s{T} \to \s{T}^\prime$ if, for any pair of concepts $C = (\s{T}, \s{M})$ and $C^\prime = (\s{T}^\prime, \s{M})$, the following diagram commutes for all objects $\omega \in \mathcal{M}$:
\vspace*{-2mm}
\[{\small 
    \begin{tikzcd}[column sep=1.3cm, row sep=.5cm]
    \s{M} \arrow[d, "\operatorname{id}"'] \arrow[r, "\gamma", bend right=18] & \s{T} \arrow[l, "\beta"', bend right=18] \arrow[d, "\tau_\s{C}"] \\
    \s{M} \arrow[r, "\gamma^\prime"', bend right=18] & \s{T}^\prime \arrow[l, "\beta^\prime", bend right=18] 
    \end{tikzcd}
}\]
\end{definition3}

\begin{example}
    Given a set of objects $\mathcal{U} = \{\inlineimg{appleb}, \inlineimg{oneblack}, \inlineimg{onewhite}\}$, the sentences $\s{T} = \{\texttt{black}\}$ and $\s{T}^\prime = \{\texttt{noir}, \texttt{un}\}$, and the objects $\s{M} = \{\inlineimg{appleb}, \inlineimg{oneblack}\} \subset \s{U}$, the translation $\tau_\s{C} = \{\texttt{black} \to \texttt{noir}\}$ is sound as it preserves concept closure, while $\tau = \{\texttt{black} \to \texttt{un}\}$ is not
    \[
    \begin{adjustbox}{max width=.9\linewidth}
        \begin{tikzcd}[column sep=1.3cm, row sep=.5cm]
        \{\inlineimg{appleb}, \inlineimg{oneblack}\} \arrow[d, "\operatorname{id}"'] \arrow[r, "\gamma", bend right=18] & \{\texttt{black}\} \arrow[l, "\beta"', bend right=18] \arrow[d, "\tau_\s{C}"] \\
        \{\inlineimg{appleb}, \inlineimg{oneblack}\} \arrow[r, "\gamma^\prime"', bend right=18] & \{\texttt{noir}\} \arrow[l, "\beta^\prime", bend right=18] 
        \end{tikzcd}
        \quad
        \begin{tikzcd}[column sep=1.3cm, row sep=.5cm]
        \{\inlineimg{appleb}, \inlineimg{oneblack}\} 
        \arrow[d, "\cancel{\operatorname{id}}"']
        \arrow[r, "\gamma", bend right=18] & \{\texttt{black}\} \arrow[l, "\beta"', bend right=18] \arrow[d, "\tau"] \\
        \{\inlineimg{onewhite}, \inlineimg{oneblack}\} \arrow[r, "\gamma^\prime"', bend right=18] & \{\texttt{un}\} \arrow[l, "\beta^\prime", bend right=18] 
        \end{tikzcd}
    \end{adjustbox}
    \]
\end{example}
Having defined concepts and sound translations, we give a probabilistic interpretation of concepts in order to characterise models for which these translations are sound.


\paragraph{Probabilistic interpretation of concepts.}
To reason about concepts probabilistically, for each concept indexed by $\cidx$, we denote as $\s{M}_{\cidx}$ the set of objects associated with that concept. We then introduce a map $g_{\cidx} : \s{X} \to \s{C}$, such that, for each representation $x \in \s{X}$, the value $g_{\cidx}(x)$ represents possible concept outcomes in $\s{C}$ associated with the $\cidx$-th concept. This allows the definition of a concept random variable as a map $\rv{C}_{\cidx} : \Omega \to \s{C}$ defined by the composition:
\vspace*{-2mm}
    \[
    \begin{tikzcd}
    \Omega
    \arrow[r, "\rv{X}"]
    \arrow[dr, "\rv{C}_{\cidx}"']
    & \s{X}
    \arrow[d, "g_{\cidx}"] \\
    {} & \s{C}
    \end{tikzcd}
    \vspace*{-2mm}
    \]
Under this view, a \textit{concept-based model} can be obtained by marginalising $\pdist_{\rv{Y} \mid \rv{Z}}$ as $\pdist_{\rv{Y} \mid \rv{C}} \circ \pdist_{\rv{C} \mid \rv{Z}}$ where $\pdist_{\rv{C} \mid \rv{Z}}: \s{Z} \to \Prob{\s{C}}$ is a \emph{concept encoder} and $\pdist_{\rv{Y} \mid \rv{C}}: \s{C} \to \Prob{\s{Y}}$ is a \emph{concept-based task predictor}. 
This marginalisation makes $\rv{Y}$ conditionally independent to $\rv{Z}$ given $\rv{C}$, which leads us to conclude that:\looseness-1

\smallskip
\begin{definition4}
Verifying inference equivariance for $\pdist_{\rv{Y} \mid \rv{C}}$ is equivalent to that for $\pdist_{\rv{Y} \mid \rv{Z}}$, but it supports the use of sound translations.
\end{definition4}

\subsubsection{Consequences of Concept Closure}

\myparagraph{Derivable properties.}
Models that support sound translations implicitly allow for their latent spaces to be \emph{aligned} with a user's vocabulary, a property advocated for in concept-based explainability methods~\citep{kim2018interpretability, cbm}. Similar properties advocated by traditional views of interpretability, such as \emph{faithfulness} and \emph{fidelity} \citep{jacovi2020towards}, emerge as consequences of the symmetry above, as a translation that is not closed over the target concept space may induce multiple, potentially conflicting interpretations of the same model behaviour.

\myparagraph{Methods to guarantee concept-closure invariance.}
We notice that existing interpretable models already implicitly advocate for a similar invariance as the one we just described: 
\textit{Concept Bottleneck Models} (CBMs) \citep{cbm} enforce concept-closure by construction by forcing the model to use concepts from the user's conceptual space.

\myparagraph{Disqualified approaches.}
Our requirement for interpretable models to satisfy concept closure formalises why it is common to think of traditional machine learning models such as decision trees \citep{breiman1984classification,hu2019optimal} or additive models \citep{hastie1986generalized,agarwal2021neural} as uninterpretable when they operate on non-concept spaces (e.g., individual pixels).

\subsection{Symmetry IV: Structural Invariance}
\label{sec:maps}
\Cref{def:sym-two} and \Cref{def:sym-three} reduce the analysis of inference equivariance to $\pdist_{\rv{Y} \mid \rv{C}}$. However, the ``behaviour'' of this model has not yet been characterised. To do this, we consider challenge \textbf{C3} and ask the following research question: (\textbf{RQ4})~\emph{How should an interpretable model behave?}

The ``behaviour'' of a model is determined by its \emph{structural properties} (such as linearity or monotonicity), which describe how outputs are computed from inputs. Knowing these properties allows a user to know what the model can and cannot do, enabling reasoning and control over its behaviour. Note that which properties matter is user dependent, as they define the user's hypothesis space $\cat{Hm}$ of ``mental models''. When $\pdist_{\rv{Y} \mid \rv{C}}$ satisfies the structural properties encoded in $\cat{Hm}$, the user can internally simulate the model, as required by inference equivariance.
\begin{example}
Consider a conditional model $P_{Y\mid X_1,X_2}:\{0,1\}^2\to\Prob{\{0,1\}}$ defined by
$P_{Y\mid X_1,X_2}(x_1,x_2)=\mathrm{Bernoulli}(\sigma(\alpha(x_1+x_2-2x_1x_2)+\beta)$, which computes a probabilistic XOR. Consider two human users: a student whose hypothesis space consists of linear-logistic kernels of the form $P_{Y\mid X_1,X_2}^{[h_1]}(x_1,x_2)=\mathrm{Bernoulli}(\sigma(w_1x_1+w_2x_2+b))$, and a researcher whose hypothesis space includes piecewise-linear kernels of the form $P_{Y\mid X_1,X_2}^{[h_2]}(x_1,x_2)=\mathrm{Bernoulli}(\sigma(\ell(x_1,x_2)))$ with $\ell(x_1,x_2)=a_1x_1+a_2x_2+a_3\max\{0,x_1+x_2-1\}+b$. Since the model $P_{Y\mid X_1,X_2}$ is not contained in the student's hypothesis class, inference equivariance fails for the student but holds for the researcher, whose hypothesis space includes the model.
\end{example}

To formalise this intuition, we introduce a functor $F$ between a category of models $\cat{Im}$ (standing for interpretable models) 
and a category of models $\cat{Hm}$ (standing for human mental models). 
Structural invariants are defined as the properties preserved under $F$, that is, the properties that both $\cat{Im}$ and $\cat{Hm}$ models have in common.

\smallskip
\begin{definition3}\textbf{(Structural Invariance)}
\label{def:sym-four}
The structural properties of models in $\cat{Im}$ are invariant under the functor $F: \cat{Im} \to \cat{Hm}$ if and only if there exist two injective functors $E_1: \cat{Im} \hookrightarrow \cat{BorelStoch}$ and $E_2: \cat{Hm} \hookrightarrow \cat{BorelStoch}$ such that the following diagram commutes:
\vspace*{-2mm}
\[
\begin{tikzcd}
\cat{Im}
  \arrow[r, hook, "E_1"]
  \arrow[d, "F"']
&
\cat{BorelStoch}
\\
\cat{Hm}
  \arrow[ru, hook, "E_2"']
&
\end{tikzcd}
\]

\end{definition3}


If interpretable models satisfy structural invariance, we can conclude the following:

\smallskip
\begin{definition4}
    If $\pdist_{\rv{Y} \mid \rv{C}}$ is compatible with a user's mental model's hypothesis space, then that user can predict the model's results and verify inference equivariance.
\end{definition4}

\subsubsection{Consequences of Structural Invariance.}

\myparagraph{Derivable properties.}
Structural invariance makes interpretability explicitly user-centric and task-specific: the structural properties required of a model depend on the user's ``mental model'' \citep{johnson1983mental,kulesza2013too,kim2018interpretability}. These properties yield transparency by exposing the model's specific hypothesis space. As a result, properties such as \emph{linearity}, \emph{monotonicity}, or \emph{sparsity} are not first principles but instantiations of structural properties chosen to match a given user.

\myparagraph{Methods to guarantee structural invariance.}
Structural invariance can be achieved by fixing the model's hypothesis space a priori. For instance, one may restrict this space to encompass only linear models or monotone functions \citep{rudin2019stop,debot2024interpretable}. More flexibility can be achieved with models that explicitly factorise concept dependencies \citep{dominici2024causal}.

\myparagraph{Disqualified approaches.}
Sparse autoencoders and concept-based models whose mappings between concepts and tasks (or between concepts themselves) are arbitrary, such as employing DNNs, have a hypothesis space that lacks any well-defined structure and therefore do not satisfy this symmetry. More broadly, models that violate domain-required structures must be excluded: for example, in medical risk scoring, where monotonicity is a critical structural requirement, highly expressive models such as unconstrained DNNs are not interpretable for that domain, regardless of their predictive accuracy.

\section{Probabilistic Composition of Interpretable Models}
\label{sec:models}
In this section, we leverage the discussed symmetries to formally answer the following question: (\textbf{RQ5})~{\emph{How can we formalise interpretable models?}}

We answer this by first giving a complete formal definition of an \emph{interpretable model} in terms of its symmetries, and then presenting the \emph{category of interpretable models}, that is, a general ``recipe'' for building interpretable models.

\smallskip
\begin{definition2}[Interpretable Models]
    Interpretable models are those that satisfy all interpretability symmetries.
\end{definition2}
The general category of interpretable models (as a subcategory of $\cat{BorelStoch}$) is specified by the objects, processes, and compositional rules from which any interpretable model can be constructed. To present this categorical structure in an intuitive yet rigorous way, we represent models using \emph{string diagrams} \citep{joyal1991geometry,selinger2010survey,baez2010physics} for Markov categories \citep{cho2019disintegration,fritz2020synthetic,jacobs2025compositional}. These diagrams depict how value spaces of random variables are connected and composed by probabilistic transformations. While offering a compelling visual intuition, string diagrams are fully formal objects, much like electrical or quantum circuits. They are read from left to right: boxes represent probabilistic transformations (playing a role analogous to layers in a DNN), while wires represent the value spaces of the input and output random variables associated to each transformation (for a formal and complete introduction, see~\citet{fritz2020synthetic}).

\smallskip
\begin{definition2}(Category of Interpretable Models)
    \label{def:category_of_interpretable_models}
    The category of interpretable models $\cat{Im} \hookrightarrow \cat{BorelStoch}$ has: 
    \begin{itemize}[topsep=0pt, leftmargin=9pt, itemsep=-2pt]
        \item \textbf{Objects}: concept spaces $\s{C}_1, \dots, \s{C}_k$.
        \item \textbf{Processes}: concept-based conditional probability distributions $\pdist_{\rv{C}_i \mid \rv{C}_j}: \s{C}_j \to \Prob{\s{C}_i}$, copy maps $\textrm{copy}_{\s{C}_i}: \s{C}_i \to \s{C}_i \times \s{C}_i$ (duplicating $\s{C}_i$), and discard maps $\textrm{discard}_{\s{C}_i}: \s{C}_i \to 1$ (marginalising, i.e., ignoring or summing out $\s{C}_i$).
        \[
            \resizebox{.9\columnwidth}{!}{\compositionalElements}
        \]
        \item \textbf{Composition rules}: sequential and parallel composition of processes.
        \[
            \resizebox{\columnwidth}{!}{\compositions}
        \]
    \end{itemize}
\end{definition2}
\begin{remark}
Note that, if $\s{Y}$ is a concept space, that is $\s{Y} \equiv \s{C}_i$, then $\pdist_{\rv{Y} \mid \rv{C}}$ is a valid process in this category. Moreover, we can enrich this category with processes involving $\s{X}$, $\s{Z}$, or parameter spaces $\Theta$ to build arbitrary complex models. Finally, note that while this category describes a general recipe to designing potentially interpretable models, a model in this category is concretely interpretable if and only if it satisfies all interpretability symmetries with respect to a reference user $h$.
\end{remark}


A key advantage of string diagrams is that they also represent probabilistic inference diagrammatically, providing a unified way to reason about inference in interpretable models. 
In the following, for readability, we accompany each diagram with an equivalent analytical expression familiar to the probabilistic machine learning community\footnote{To support Bayesian inference, hereafter, we use string diagrams for \emph{partial} Markov categories \citep{di2025partial} which has to be interpreted in the category $\cat{BorelStoch_{\leq1}}$ of sub-Markov kernels \citep{panangaden1999category,di2025partial}.}.

\section{Inference on Probabilistic Interpretable Models}
\label{sec:inference}
Now that we formalised interpretable models, we proceed to describe inferences on such models, that is, how interpretable models can be used in practice. We will describe three of the most common and widely used types of inference used in interpretability: concept alignment (Sec~\ref{sec:alignment}), interventions, and counterfactuals (Sec~\ref{sec:interaction}).

\subsection{Concept Alignment}
\label{sec:alignment}
Here, we discuss how the inference process allows the model to align its concepts w.r.t.\ a target user. We thus ask the question: (\textbf{RQ6})~{\emph{How can we learn human concepts?}}

The process of matching a model's concepts to human concepts is called \emph{alignment}. In a nutshell, we aim to compute the distribution of parameters $\Theta$ that make the prediction of a concept map match a set of ground-truth concepts $c$. From a probabilistic perspective, this corresponds to computing the posterior distribution over the parameters $\Theta$ given a set of ground-truth concepts $c$ as evidence. In diagrammatic terms, we can represent observing the evidence $\rv{C}=c$ by bending the $\s{C}$ wire backwards and inserting a constant state representing the evidence $c$. We indicate the normalisation integral in the Bayesian posterior with a blue dashed box and use $\textrm{pa}(\s{C})$ to refer to the inputs of a model outputting $\s{C}$.

\smallskip
\begin{definition2}[Concept Alignment]
Given a parametric concept map $\pdist_{\rv{C} \mid \textrm{pa}(\rv{C}); \Theta}$, concept alignment is the Bayesian inversion represented by the diagram:
\begin{gather*}
    \resizebox{.8\columnwidth}{!}{\conceptAlignment}\\
    \resizebox{1.\columnwidth}{!}{$\pdist_{\Theta \mid \rv{C}=c, \text{pa}(\rv{C})}(\theta) = \frac{\pdist_{\rv{C} \mid \text{pa}(\rv{C}); \Theta}(c \mid \text{pa}(C); \theta) \pdist_{\Theta}(\theta)}{\int_\Theta \pdist_{\rv{C} \mid \text{pa}(\rv{C}); \Theta}(c \mid \text{pa}(C); \theta) \pdist_{\Theta}(\theta) \text{d}\theta}$}
\end{gather*}
\end{definition2}

\subsection{Human-machine interaction}
\label{sec:interaction}
Given an aligned interpretable model, we explore how users may interact with it. In particular, we ask: (\textbf{RQ7})~{\emph{Which queries are supported by interpretable models?}}

\paragraph{Interventional inference.}
A key advantage of concept-based models is their support for human interaction, allowing users to \textit{intervene} and change concept activations in various ways \cite{cbm, coop, barker2023selective, closer_look_at_interventions, collins2023human, intcem, beyond_cbms}. In general, interventions can be described as a form of posterior inference where we consider a concept-based process $\pdist_{\rv{C} \mid \text{pa}(C)}$ as a \emph{prior}, and introduce a \emph{likelihood} $\pdist_{\text{pa}'(\rv{C}) \mid \rv{C}}$ to update our prior after observing random variables $\text{pa}'(\rv{C})$:

\smallskip
\begin{definition2}[Intervention]
Given a parametric concept map $\pdist_{\rv{C} \mid \textrm{pa}(\rv{C})}$ and a likelihood $\pdist_{\textrm{pa}'(\rv{C}) \mid \rv{C}}$, an intervention is the Bayesian inversion represented as:
\vspace*{-3mm}
\begin{gather*}
    \resizebox{.8\columnwidth}{!}{\interventionPosterior}\\
    \resizebox{1.\columnwidth}{!}{$\pdist_{C \mid \mathrm{pa}(C), \mathrm{pa}'(C)}(c \mid \mathrm{pa}(c), \mathrm{pa}'(c)) = 
    \frac{\pdist_{\mathrm{pa}'(C)\mid C}(\mathrm{pa}'(c)\mid c)\,
    \pdist_{C\mid \mathrm{pa}(C)}(c\mid \mathrm{pa}(c))}
    {\int_{C}
    \pdist_{\mathrm{pa}'(C)\mid C}(\mathrm{pa}'(c)\mid c)\,
    \pdist_{C\mid \mathrm{pa}(C)}(c\mid \mathrm{pa}(c))
    \,\mathrm{d}c}$}
\end{gather*}
\end{definition2}
Under this lens, \textit{ground-truth interventions}~\citep{cbm}, seen in left image below and represented as $\pdist_{\rv{C} \mid gt(\textrm{pa}(\rv{C}), \textrm{pa}'(\rv{C}))} = \pdist_{\rv{C} \mid \textrm{pa}'(\rv{C}) = c^*}$, become a special case where the prior becomes a uniform distribution independent on $\text{pa}(C)$ (which gets discarded), the likelihood is the identity function, and the evidence is represented by ground-truth labels $c^*$. In contrast, \textit{do-interventions}~\cite{pearl1988probabilistic}, seen in the right image below and represented as $\pdist_{\rv{C} \mid do(\textrm{pa}(\rv{C}), \textrm{pa}'(\rv{C}))} = \pdist_{\rv{C} \mid \textrm{pa}'(\rv{C}) = k}$ , are a special case where the evidence is a constant value $k \in \mathbb{R}$:
\begin{gather*}
    \resizebox{.45\columnwidth}{!}{\groundTruthIntervention}
    \resizebox{.45\columnwidth}{!}{\doIntervention}\\
\resizebox{1.\columnwidth}{!}{
$\pdist_{\rv{C} \mid \mathrm{pa}(\rv{C}),\,\mathrm{pa}'(\rv{C})}
(c \mid \mathrm{pa}(c), c^*) = \mathbb{I}_{\{c=c^*\}} \quad
\pdist_{\rv{C} \mid \mathrm{pa}(\rv{C}),\,\mathrm{pa}'(\rv{C})}
(c \mid \mathrm{pa}(c), k) = \mathbb{I}_{\{c=k\}}$
}
\end{gather*}

\paragraph{Counterfactual inference.}
Counterfactual inference is a cornerstone of both interpretable and causal machine learning, enabling the evaluation of causal effects in hypothetical scenarios. Following the standard causality approach \citep{jacobs2025compositional}, counterfactual inference requires rewriting a given model in such a way that all randomness is confined in a set of source distributions $\pdist_{U_i}$ referred to as ``exogenous'' variables that capture high-dimensional and potentially entangled information. Therefore, within our framework counterfactual can be expressed as follows:


\begin{definition2}[Counterfactual]
    Given a model $\pdist_{\rv{Y} \mid \rv{C}_1,\dots,\rv{C}_n} \circ \pdist_{\rv{C}_1,\dots,\rv{C}_n \mid \rv{U}} \circ \pdist_{\rv{U}}$, a counterfactual is the following sequence of Bayesian inversions:
    \begin{itemize}[topsep=0pt, leftmargin=9pt, itemsep=-2pt]
        \item \textbf{Abduction:} observe the value of target concepts $\rv{Y}$ to compute the posterior over exogenous variables:
        \resizebox{.8\columnwidth}{!}{\counterfactualInferenceAbduction}
        \item \textbf{Action:} given the posterior on $\rv{U}$, duplicate the model and intervene on the new model:\\\
        \resizebox{.8\columnwidth}{!}{\counterfactualInferenceAction}
        \item \textbf{Prediction:} infer the value of $\rv{Y}$ in the duplicated intervened model $\pdist_{\rv{Y} \mid do(\rv{C})}$.
    \end{itemize}
\end{definition2}

Taken together, the previous three definitions allow us to conclude the following:

\begin{definition4}
    Alignment, interventional, and counterfactual inference are all forms of Bayesian inversion.
\end{definition4}

\subsection{Consequences of Probabilistic Interpretability}


The probabilistic framework of interpretability described above allows us to derive the following consequences:

\myparagraph{Derivable properties.}
Model actionability comes as a consequence of inferences on interpretable models: alignment, interventions, and counterfactuals have actionability \citep{poyiadzi2020face} and interactivity \citep{tenney2020language} as direct consequences. Moreover, alignment can be formally measured within this framework, inheriting the same theoretical guarantees and generalisation bounds as probabilistic inference itself.

\myparagraph{Methods for probabilistic interpretable inference.}
Unifying alignment, interventional, and counterfactual inference means that a single posterior inference algorithm, such as belief propagation, can, in principle, be employed to perform all interpretable inferences, providing a coherent and computationally grounded foundation for probabilistic interpretable inference.

\section{Alternative Views}
\label{sec:alternative}
In this section we discuss alternative views of interpretability. For each view, we then provide a rebuttal in terms of the symmetries and definitions we proposed in this work.

\paragraph{\textcolor{mediumred}{[Alternative position \#1]} There is no universal, mathematical definition of interpretability, and there never will be ~\citep{murphy2023probabilistic}}
As discussed, most works in the interpretability literature define their subject matter informally. As a consequence, the literature has gravitated toward enumerating extensive lists of desirable properties (modularity, simplicity, stability, completeness, actionability, etc.) rather than converging on unifying principles. Moreover, many authors observe how interpretability is necessarily user-centric and task-specific, leaving to some, such as~\citet{murphy2023probabilistic}, to conclude that \emph{there is no universal, mathematical definition of interpretability}.\looseness-1\\
\textcolor{mediumblue}{\textbf{Rebuttal:}} While this view reflects the current state of the field, the absence of formal grounding has generated significant ambiguity, undermining the coherence and progress of interpretable AI research. Our position resolves this fragmentation by identifying a small set of general formal principles from which the properties discussed in the literature can be derived. Importantly, Symmetries~I,~III, and~IV explicitly incorporate a human reference, thereby accounting for user-centrism and task-dependence without abandoning mathematical universality.

\paragraph{\textcolor{mediumred}{[Alternative position \#2]} Symmetries are not enough to define interpretability}
A potential objection is that symmetries alone may be insufficient to capture all essential aspects of interpretability. In particular, several properties commonly associated with interpretability do not yet have an agreed-upon formalisation in the literature (e.g., there are many different perspectives on what ``simple'' is), making it difficult at present to establish whether symmetries are adequate to account for them. Moreover, interpretable AI remains a relatively young field, and future research may uncover additional properties that require either further symmetries or an altogether different formal framework.\\
\textcolor{mediumblue}{\textbf{Rebuttal:}} The introduced symmetries are designed to capture existing properties that have already been discussed in the literature, and we explicitly indicate how specific interpretability properties might be derived as consequences of specific symmetries. In this sense, the framework is extensible: newly identified properties can be analysed to determine whether they correspond to additional symmetries or motivate refinements of the existing ones.

\paragraph{\textcolor{mediumred}{[Alternative position \#3]} Symmetries are not suitable for the interpretability community.}
Even if symmetries are expressive enough in principle, a further concern is that the proposed symmetries are formulated using mathematical notions that may be unfamiliar to many researchers in the interpretability community. This could create a barrier to understanding and adoption, undermining their practical value as a foundation for interpretability.\\
\textcolor{mediumblue}{\textbf{Rebuttal:}} There is an inherent trade-off between formal rigour, which enables clarity and generality, and accessibility, which lowers barriers to entry. Our formalisation aims to strike a balance between these competing goals: while mathematically precise, each symmetry admits intuitive, example-level interpretations that can be understood without full mastery of the underlying formalism. This allows a broad audience to develop an operational intuition for the proposed first principles, without sacrificing the generality and unifying power that formalisation affords.


\section{Conclusion}
To realise the aims of this position, the community should prioritise a systematic formal validation of the proposed first principles by showing that existing interpretability desiderata and evaluation criteria can be derived as consequences of the proposed symmetries. This requires researchers explicitly stating which symmetry assumptions their methods rely on and which interpretability properties follow from them, thereby clarifying when a method is complete with respect to the above symmetries.\looseness-1

As a concrete example, the concept-based interpretability community has largely focused on concept-closure invariance~\citep{poeta_ciravegna_concept_learning_survey}, but has paid comparatively little attention to structural invariance. As a result, concept-based models often exhibit arbitrary behaviour (e.g., when using DNNs as task predictors) or are weak classifiers (e.g., when using linear models as task predictors). Conversely, the mechanistic interpretability community has been less attentive to concept-closure invariance and information invariance. This typically leads to concepts that are not aligned with human semantics and to information overload.

We therefore call on researchers to (i)~map existing interpretability methods to the proposed symmetries, (ii)~identify which symmetries are implicitly assumed or violated in their work, and (iii)~develop new methods that are explicitly symmetry-complete w.r.t.\ their intended tasks and users.

\bibliography{example_paper}
\bibliographystyle{icml2025}

\end{document}